\documentclass[twocolumn,10pt]{article}

\usepackage{booktabs}   
\usepackage{subcaption} 
\usepackage{url}
\usepackage{amsmath}
\usepackage{latexsym}
\usepackage[pdftex]{graphicx}  
\usepackage{authblk}

\title{Profiling based Out-of-core Hybrid Method \\
for Large Neural Networks}

\author[1]{Yuki Ito \footnote{Currently, Yahoo! JAPAN Corporation}}
\author[2]{Haruki Imai}
\author[2]{Tung Le Duc}
\author[2]{Yasushi Negishi}
\author[2]{\\Kiyokuni Kawachiya}
\author[1]{Ryo Matsumiya \footnote{Currently, DeNA Co., Ltd.}}
\author[1]{Toshio Endo}

\affil[1]{Tokyo Institute of Technology}
\affil[2]{IBM Research - Tokyo}

\date{}

\begin{document}

\maketitle

\begin{abstract}
GPUs are widely used to accelerate deep learning with NNs (NNs). On the other hand, since GPU memory capacity is limited, it is difficult to implement efficient programs that compute large NNs on GPU. To compute NNs exceeding GPU memory capacity, data-swapping method and recomputing method have been proposed in existing work. However, in these methods, performance overhead occurs due to data movement or increase of computation. In order to reduce the overhead, it is important to consider characteristics of each layer such as sizes and cost for recomputation. Based on this direction, we proposed Profiling based out-of-core Hybrid method (PoocH). PoocH determines target layers of swapping or recomputing based on runtime profiling. We implemented PoocH by extending a deep learning framework, Chainer, and we evaluated its performance. With PoocH, we successfully computed an NN requiring 50 GB memory on a single GPU with 16 GB memory. Compared with in-core cases, performance degradation was 38 \% on x86 machine and 28 \% on POWER9 machine. 
\end{abstract}

\section{INTRODUCTION}
Recently, neural networks (NNs) have archived high accuracy in many fields of machine learning such as image recognition, speech recognition, and natural language processing \cite{image_classification} \cite{dcgan} \cite{speech_classification} \cite{language_processing}. Neural networks are computation model originally based on human brain's learning mechanisms.

Since computations of NNs, especially for deep networks, are heavy tasks, GPUs have been widely used to accelerate them. However, the problem sizes of NNs that can be computed are limited by GPU memory capacity. For example, ResNet50 network \cite{resnet} with the batch size of 640 requires 50 GB memory. Even the latest GPU, NVIDIA Tesla V100, has only 16 GB or 32 GB memory, thus it is hard to support such cases on a GPU. An approach is to use multiple GPUs by harnessing data parallelism in mini-batched learning. However, in the case of 3D image learning\cite{3dnn}, memory requirement may reach dozen GBs or more even with the batch size of 1. For such cases, a different approach is required.

This paper focuses on ``out-of-core'' methods in order to exceed GPU memory capacity. There are two well known methods, data-swapping method \cite{vdnn} \cite{ooctensorflow} \cite{oocchainer} and recomputing method \cite{gradientcheckpoint}. To support larger computations, both of them suffer from performance overhead in different ways. The former suffers from CPU-GPU communication costs and the latter introduces increase in computation amounts. The better choice depends on characteristics of each NN layer, such as data size and computation type. Thus the ``hybrid'' approach is promising, where we apply either data-swapping or recomputing to NN layers. However, the optimum choice is not trivial, since we need to consider the entire NN structure, characteristics of each layer, and also the execution environment. For example, the overhead of data-swapping on a computer with PCI Express3.0 (the CPU-GPU bandwidth is 16 GB/sec) is larger than a computer with NVLINK2.0 (75 GB/sec). Therefore, the optimal swap/recompute targets are different in these two environments.

In this paper, we propose the \underline{P}rofiling-based \underline{o}ut-\underline{o}f-\underline{c}ore \underline{H}ybrid method (PoocH) for accelerating the computation of large NNs. The objective of PoocH is to reduce the performance overhead by optimizing targets of swapping and recomputing based on runtime profiling. If memory capacity allows, some layers may be kept on GPU memory with no performance overhead. Also we improve the scheduling of data movement in swapping.

We implemented PoocH by extending Chainer \cite{chainer}, a deep learning framework. The performance evaluation is done using larger NNs on two different machines, an Intel Xeon based x86 machine and a Power9 machine, each of which is equipped with a single V100 GPU with 16GB memory. The results show our hybrid method successfully reduce overheads; the performance degradation with the ResNet50 network compared with in-core cases is 38 \% on x86 machine and 28 \% on POWER9 machine.

\section{BACKGROUND}
\subsection{Neural Network}
\begin{figure}[tb]
    \centering
    \includegraphics[scale=0.23,pagebox=cropbox]{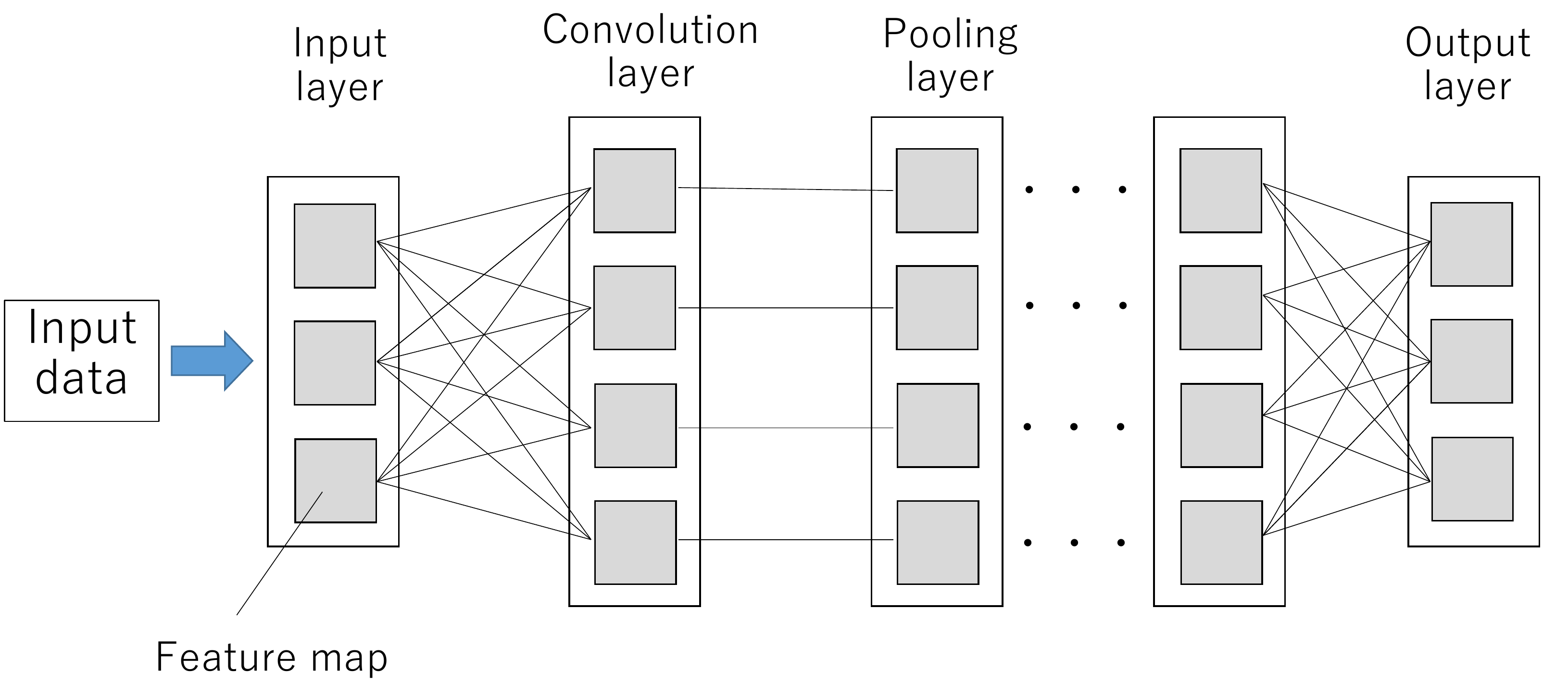}
    \caption{Structure of NN} 
    \label{fig:cnn}
\end{figure}

Figure \ref{fig:cnn} shows an example of the structure of NN. An NN is composed of multiple layers, each of which is composed of multiple feature maps. Feature maps of a layer are computed from the previous layer's feature maps.

According to the computation types, each layer is categorized into several groups: convolutional layer, pooling layer, Batch-Normalization (BN) layer, fully-connected layer, and so on. In some kinds of layer such as convolutional layer, parameters including {\em weight filters} are required for computation. Generally, the objective of learning process of NN is to find optimal values for these parameters.

The back-propagation method is a common learning method of NN, which updates parameters with values derived from errors between NN's output and correct output. In this algorithm, the following processes are performed repeatedly.
\begin{enumerate}
    \item Forward propagation: Feature maps of each layer are computed with training data as network input. The computation goes from left to right in the figure.
    \item Backward propagation: Gradients for feature maps, weight filter, etc. are computed using the errors between the output of the forward step and the correct output. The computation goes from right to left.
    \item Update: NN's parameters are updated using the gradients.
\end{enumerate}

\begin{figure}[tb]
    \centering
    \includegraphics[scale=0.25]{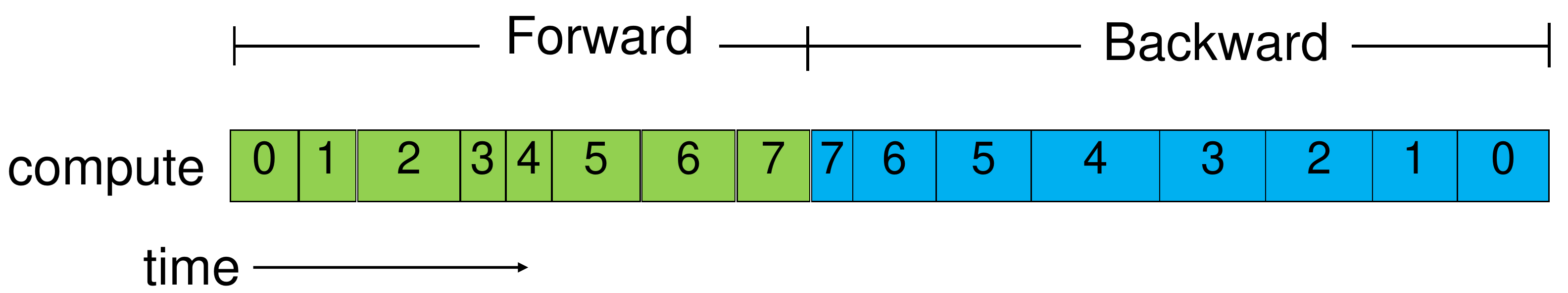}
    \captionsetup{width=0.8 \textwidth}
    \caption{Timeline of computation of NN (\#layers=8)}
    \label{fig:NN-timeline}
\end{figure}

A sample computation timeline of an NN is drawn in Figure \ref{fig:NN-timeline}. The length of each box represents the computation time of a layer. The green and blue boxes indicate forward-propagation and backward-propagation respectively. The integer in each box indicates the order of forward-propagation. Since update computation is very short usually, it is omitted in the figure. As in the figure, each propagation is computed layer by layer. While forward-propagation is conducted from input layer to output layer, back-propagation is from output layer to input layer.

Computation of backward-propagation of layer $i$ requires the feature maps, which has been computed in forward propagation of layer $i$. Thus, feature maps need to be preserved during the period.

In NN computation, ``batch processing'' technique is popular in order to utilize the degree of parallelism. With this technique, several input data (for example several images) are simultaneously given as input of the above algorithm. The number of input data is called ``batch size''.

\subsection{Memory usage of NN}
As described above, in the back-propagation method, feature maps should be preserved until they are consumed by backward computation. Also parameters such as weight filter and gradients require memory. Additionally, some computation layers may require additional workspace to accelerate the computation. 

Among them, the feature maps consume major parts. The total usage is determined by the number of layers. And the memory usage per layer increases in proportion to the batch size and the feature map size.

\begin{figure}[tb]
    \centering
    \includegraphics[scale=0.26]{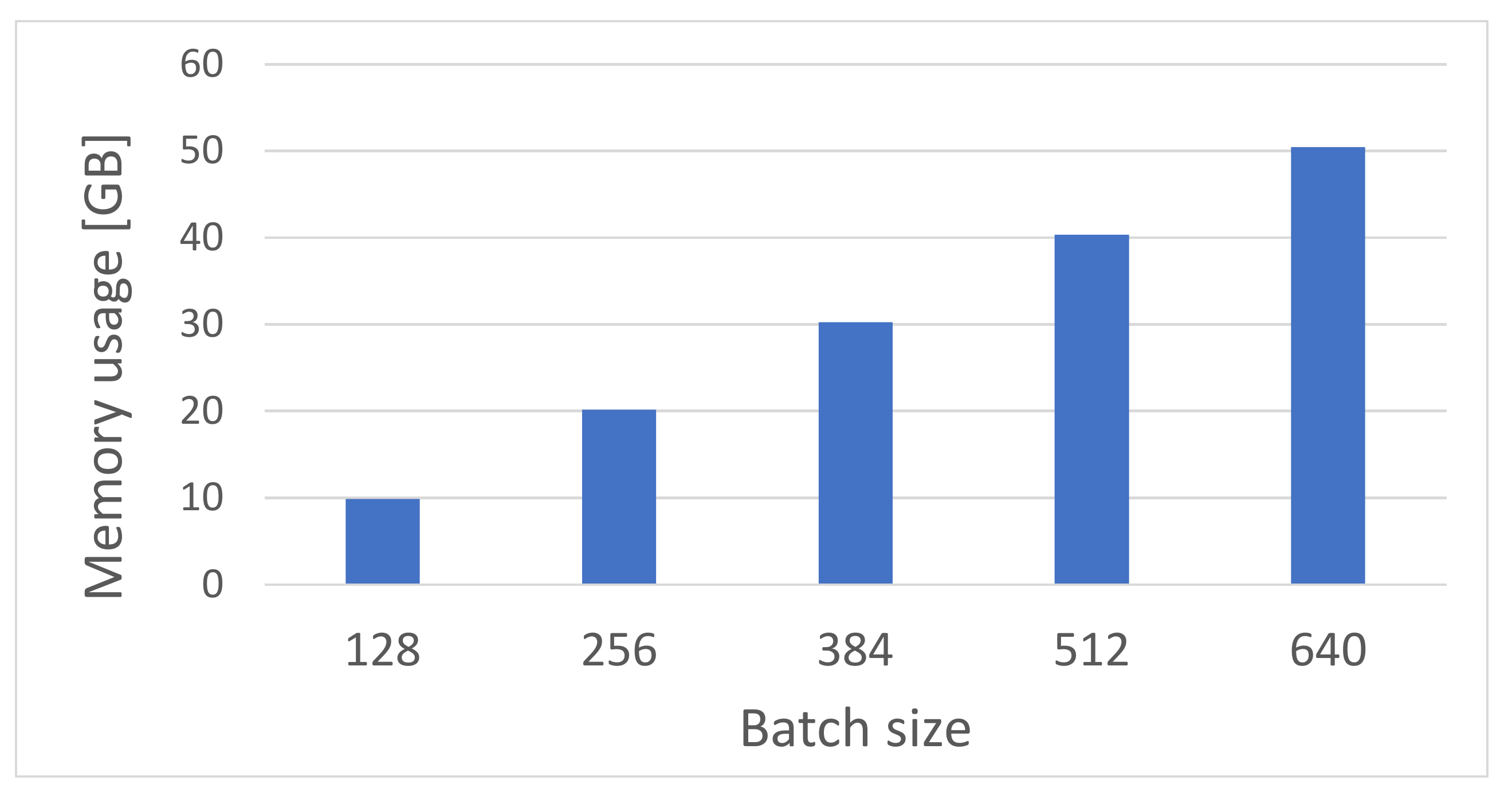}
    \caption{Memory usage of ResNet50} 
    \label{fig:resnet_memory}
\end{figure}

\begin{figure}[tb]
    \centering
    \includegraphics[scale=0.26]{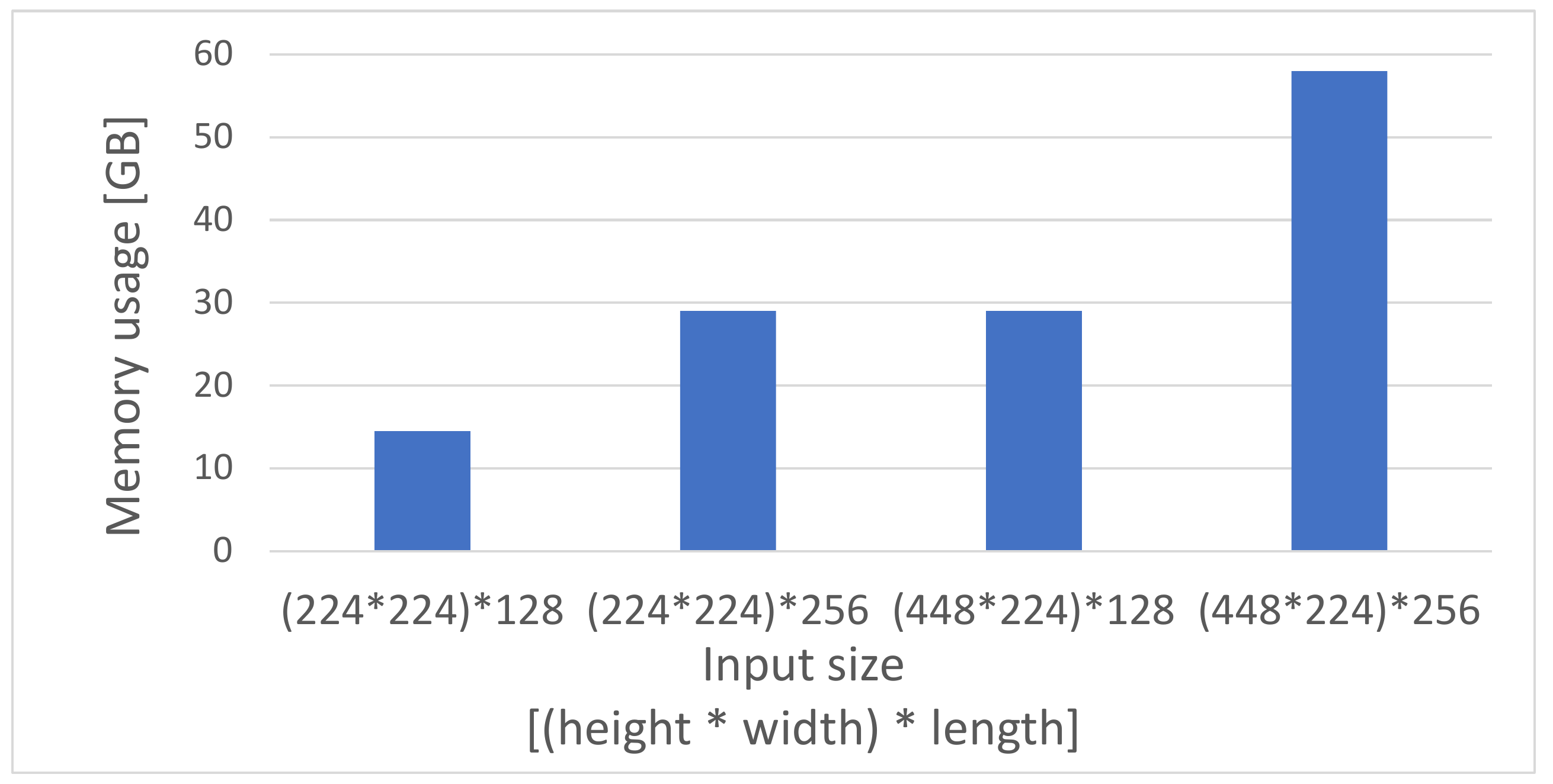}
    \caption{Memory usage of ResNext101 for 3D data (batch size = 1)} 
    \label{fig:resnext_3d_memory}
\end{figure}

The computation would be faster if all of them are allocated on GPU memory. However, large NN may require larger memory then GPU memory capacity. For example, the memory usage for computation of ResNet50 network is shown in Figure \ref{fig:resnet_memory}. The memory usage is proportional to bath size and exceeds 50 GB with the batch size of 640. As another example, the memory usage of ResNext101 for 3D data \cite{resnext} \cite{3dnn} with batch size of 1 is shown in Figure \ref{fig:resnext_3d_memory}. Memory usage increases in proportion to input data size, and reaches 58 GB in the rightmost case.

On the other hand, the memory capacity of Tesla V100 GPU we used is 16GB and it is 32 GB even on the latest product. Thus we require ``out-of-core'' execution method to support the large NNs on a single GPU.

\section{Methods to compute large scale NN}
In order to deal with data exceeding GPU memory capacity in deep learning, data-swapping method and recomputing method have been proposed. 

\subsection{Data-swapping method}
\begin{figure}[tb]
    \centering
    \includegraphics[scale=0.35]{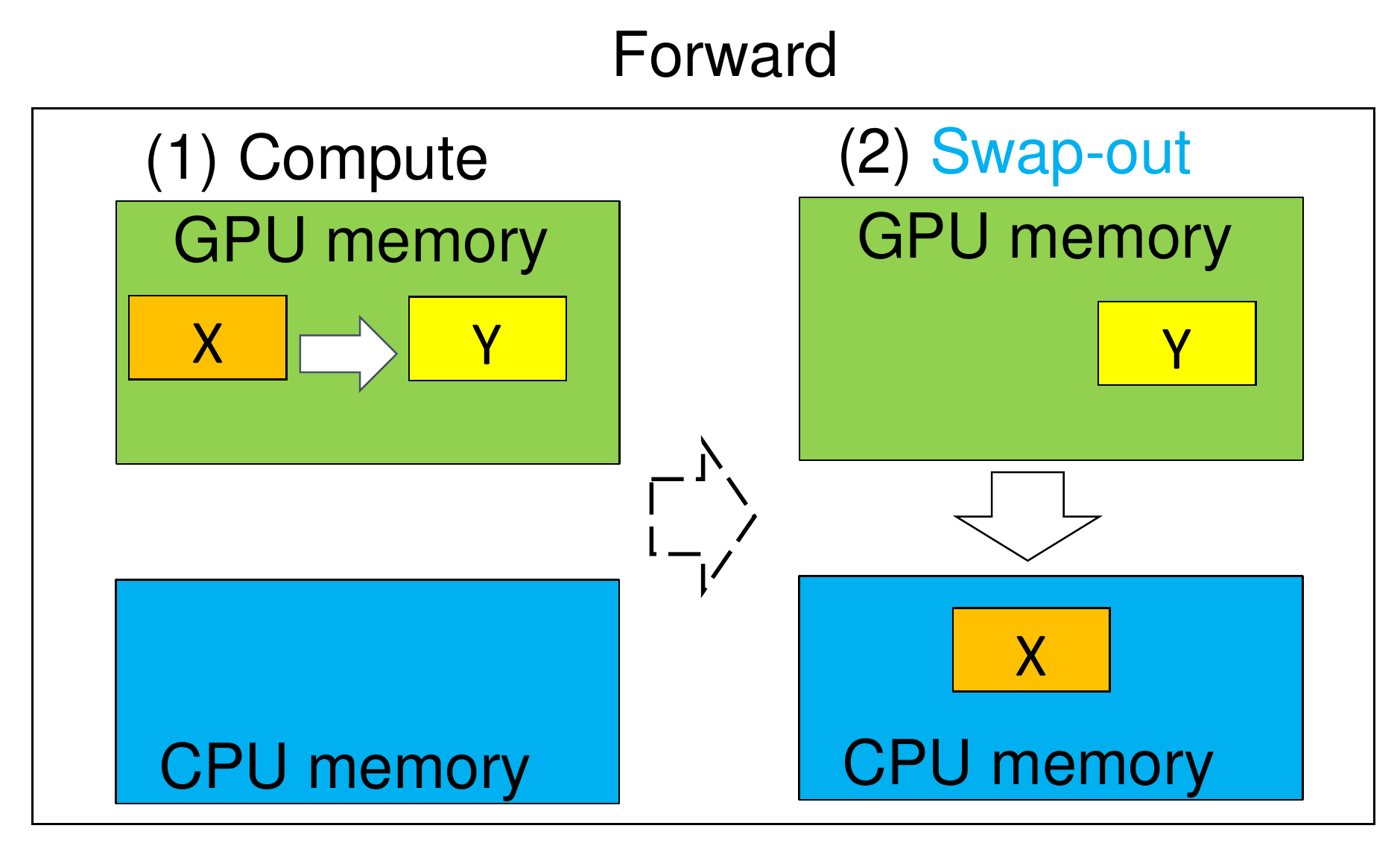}
    \caption{Swap-out at forward in data-swapping method}
    \label{fig:swapout}
\end{figure}

\begin{figure}[tb]
    \centering
    \includegraphics[scale=0.35]{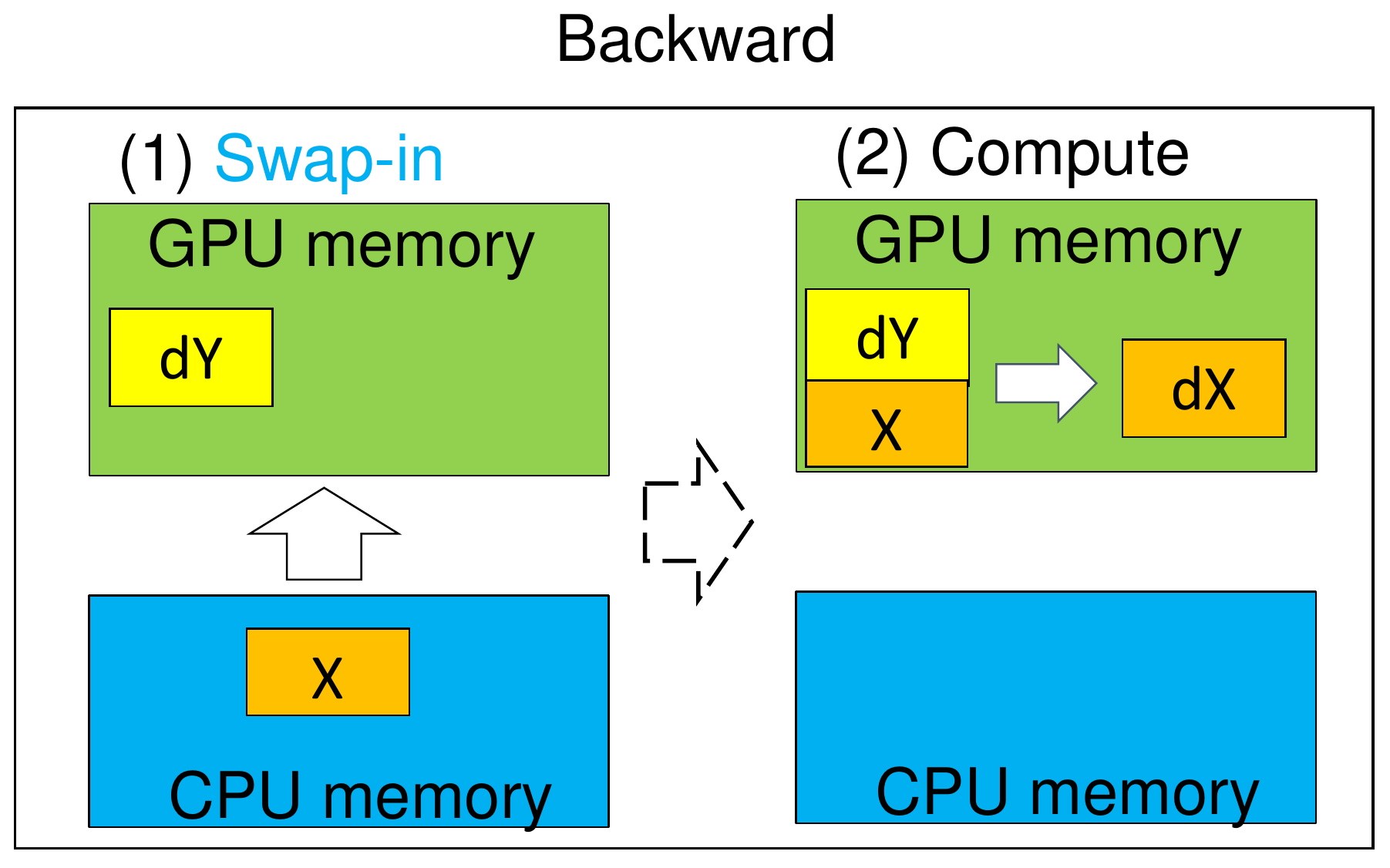}
    \caption{Swap-in at backward in data-swapping method (dX and dY are gradients corresponding to X, Y)}
    \label{fig:swapin}
\end{figure}

In the data-swapping method, a part of the data used in the forward computation of a layer is swapped out to the CPU memory. An example of swap-out is shown in Figure \ref{fig:swapout}. Here forward step of a layer computes data $Y$ from data $X$ on a GPU. After that (mainly due to memory shortage) $X$ is swapped out to CPU memory. We cannot simply discard $X$ since it is necessary in the later backward computation as shown in Figure \ref{fig:swapin}. Prior to the backward computation of this layer, $X$ on CPU memory is copied to GPU memory (swapping-in).

\subsubsection*{Impact on performance}

This method raises overhead for copying data between GPU and CPU. This overhead can be reduced by a well-known technique, overlapping of communication and computation. However, the overhead is not well hidden for the following reasons. 

The computations and swapping-in/out have to be done while keeping data dependency. Swapping-out for certain data must wait for all the forward computations that use the data. Likewise, backward computation of each layer must wait for all the necessary data to be swapped in.

\begin{figure*}[h!]
    \centering
    \includegraphics[scale=0.35]{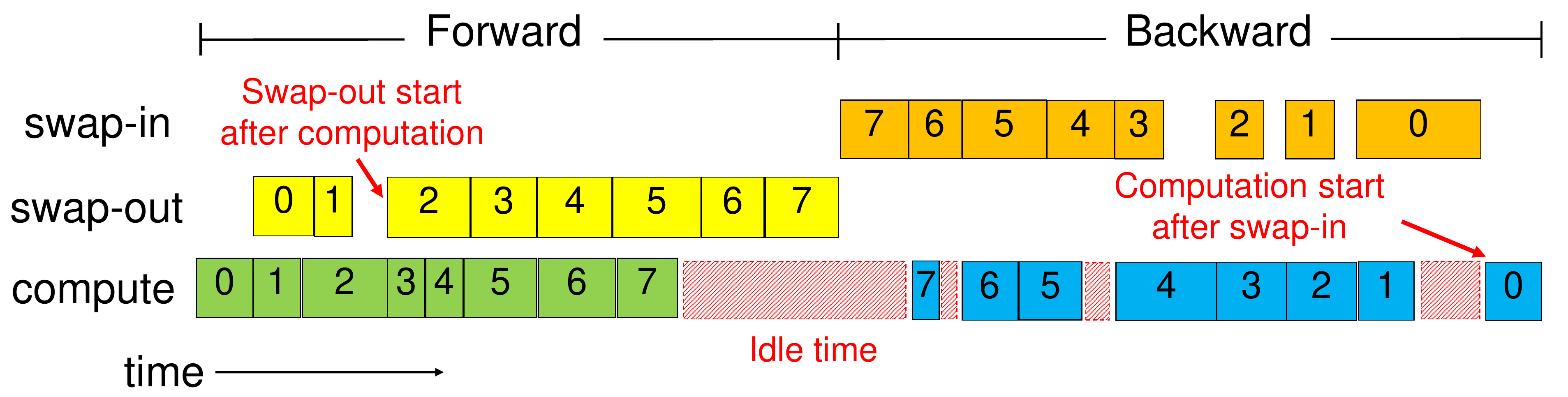}
    \captionsetup{width=0.7 \textwidth}
    \caption{Example of the timeline of NN computation processing with data-swapping (\#layers=8). The yellow and orange boxes represent the execution time of swaps corresponding to each layer's computation. In this example, data-swapping is performed in all layers and each swap-in starts simultaneously with the computation of the next layer.}
    \label{fig:swap-timeline}
\end{figure*}

Figure \ref{fig:swap-timeline} shows an example of the timeline of NN computation, where data dependency is considered. We observe several idle time regions, shown in red boxes in the figure. The largest idle region would be eliminated simply by keeping data of layers 6, 7 on GPU without swapping, if the memory capacity allows. However, there are still other idle regions, which happen when computation time is too short to hide communication costs.

\subsection{Recomputing method}
The recomputing method takes another approach to deal with small memory capacity. Some parts of the data computed in forward is simply discard without being stored anywhere. When the discarded data is required for backward computation, the data is reproduced by performing forward computation again.

\begin{figure}[tb]
    \centering
    \includegraphics[scale=0.35]{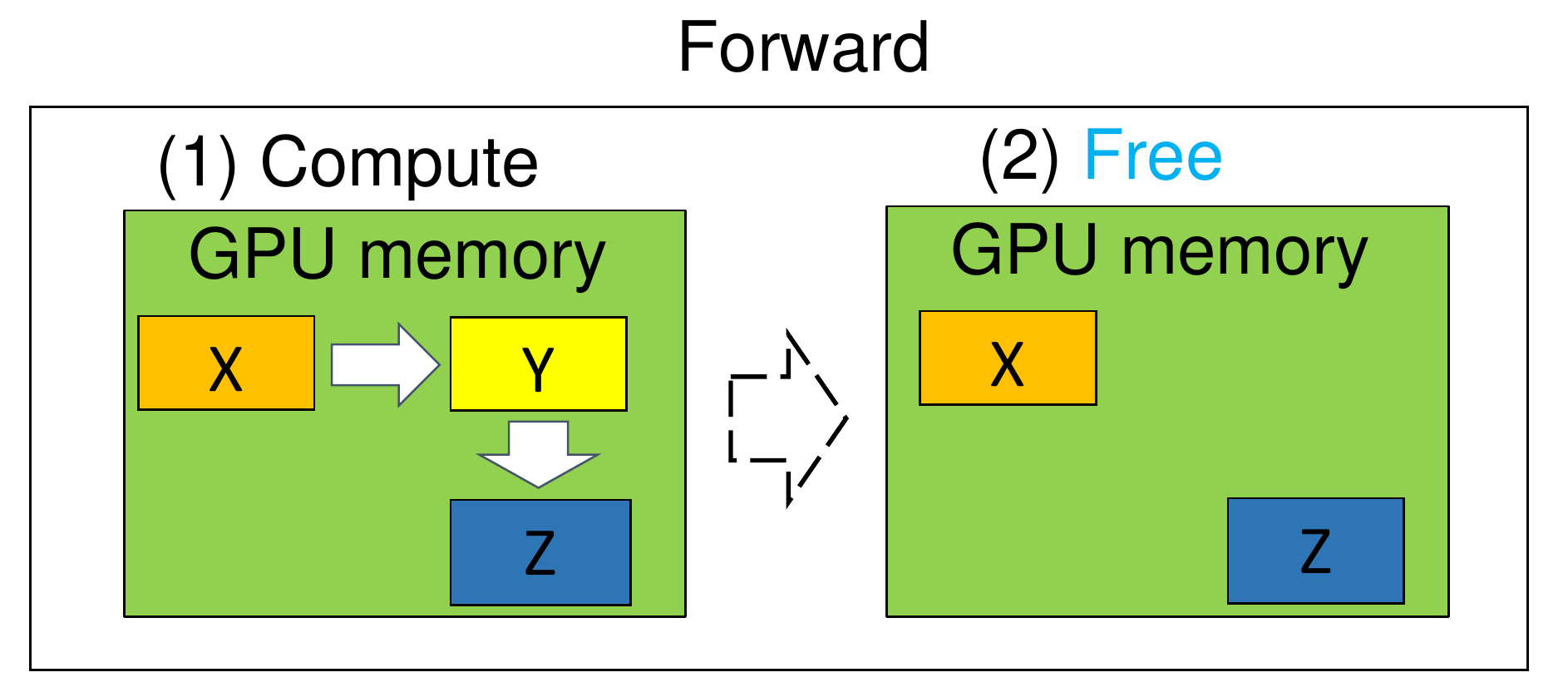}
    \caption{Free memory at forward in recomputing method}
    \label{fig:recompute-free}
\end{figure}

\begin{figure}[tb]
    \centering
    \includegraphics[scale=0.35]{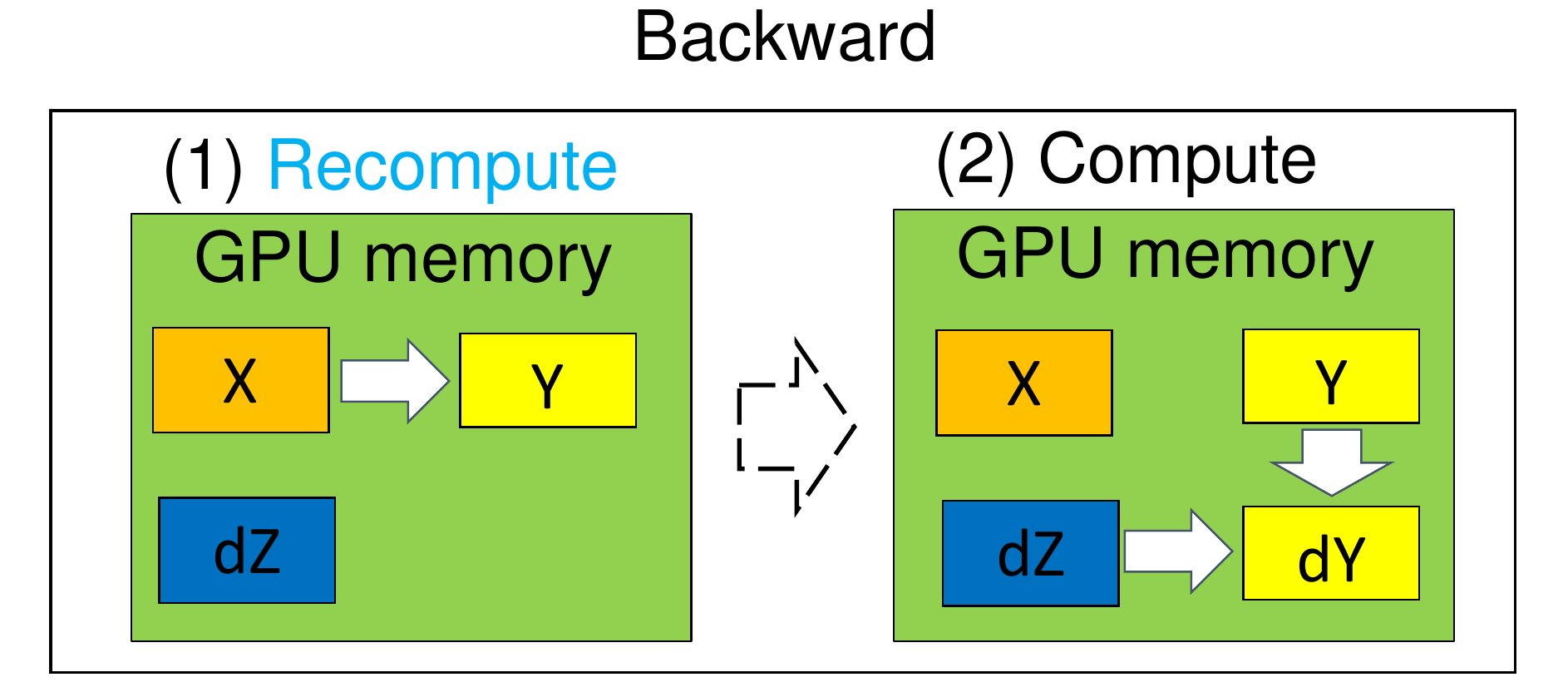}
    \caption{Recomputation at backward in recomputing method (dY and dZ are gradients corresponding to Y, Z)}
    \label{fig:recompute-recompute}
\end{figure}

An example is shown in Figures \ref{fig:recompute-free} and \ref{fig:recompute-recompute}. Here Y is computed from X, and Z is computed from Y in forward computation. After that, the data Y is discarded. After that, when Y is needed in backward (Figure \ref{fig:recompute-recompute}), Y is recomputed by performing forward computation by using X. If not only Y but X have been discarded, we would need recomputation recursively by using predecessor of X. This method suffers not from communication costs but computation costs for recomputing.

\subsection{Hybrid method}
\label{sec:hybrid}
By using one of the two methods, it is possible to compute the large scale NNs. However, data-swapping method causes CPU-GPU communication overhead, and recomputing method causes overhead due to increase of computation. The overheads have the following features, respectively.

\begin{itemize}
    \item Data-swapping method: If NN includes many layers with large computation complexity (such as convolution layer), the computation can hide the communication. If layers with small computation complexity (such as BN layer) are majority, overhead reduction is difficult.
    \item Recomputing method: If NN includes many layers with small computation complexity, the output data of the layer can be easily reproduced. For layers with large computation complexity, recomputation overhead is large.
\end{itemize}

We observe that the two methods can complement each other. Hence, utilizing both methods is promising. In the later of this paper, we call this approach ``hybrid'' methods.

One of hybrid methods is adopted in SuperNeurons \cite{superneurons}. While this method uses both swapping and recomputing, the decision criteria is very simple and determined by computation types of the target layer. For example, output of convolution layers are swapped out, and output of BN layers are recomputed. This does not take into account the actual execution time and memory usage during NN computation. On the other hand, the computation complexity and memory usage of each layer differs depending on the structure of NNs. Therefore, we expect that the performance will be improved by deciding targets of swapping and recomputing with considering the quantitative characteristics of NNs and execution environments.

\section{POOCH} 
\label{sec:pooch}
We propose Profiling-based out-of-core Hybrid method (PoocH) to accelerate the computation of large scale NNs. The key to reduce the performance overhead is to optimize targets of swapping and recomputing based on runtime profiling.

\subsection{Overview of PoocH}
\subsubsection{Optimization target}
\label{sec:target}
In PoocH, each data structure used in the computation of NN falls into one of following classes:
\begin{itemize}
    \item \textit{keep} : The data resides on GPU memory.
    \item \textit{swap} : The data is swapped out to CPU memory. When it is required later, swapped in to GPU memory.
    \item \textit{recompute} : The data is discarded. When it is required, it is recomputed.
\end{itemize}
The objective of PoocH is to find a good classification for data structures in a given NN, which realizes an execution with a short execution time, while excluding out-of-memory errors.

In the current PoocH, the classification is performed only to the feature maps of NN layers. Other data such as weight filters and gradients are retained in the GPU memory. This is because the parameters such as the weight filters tend to be much smaller than feature maps, and the lifetimes of gradient data tend to be short.

\subsubsection{Methodology of PoocH}
\label{sec:mokuteki}

The purpose of the optimization in PoocH is to reduce ``the execution time of the whole computation of NN'' while satisfying a constraint ``GPU memory usage does not always exceed the GPU memory capacity'' during the execution. In order to perform the optimization process, PoocH needs to makes suggestions of classifications repeatedly and evaluate the execution time and maximum memory usage given by each classification. However, it is difficult to formulate the execution time with a simple linear equation due to pipelined processing and data dependency. 

Instead, when a classification is chosen, PoocH simulates an execution timeline and memory management processes. By this simulation, the execution time and memory usage of the entire process can be predicted. In order to achieve this simulation, PoocH collects necessary information beforehand including computation time, swapping time of each layer by using runtime profiling.  To sum up, the flow of PoocH is as follows.

\begin{enumerate}
    \item Profiling: PoocH runs learning iterations for several times while recording necessary runtime information of each layer
    \item Classification: PoocH decides a classification of all feature maps that achieves minimal runtime
    \item Execution: The system continues learning iterations based on the classification that have chosen
\end{enumerate}

The rest of this chapter describes the profiling phase and the classification phase. Additionally, we explain additional improvements in overlapped execution in Section\ref{sec:scheduling}.

\subsection{Runtime profiling}
\label{sec:profiling}
The first step of PoocH is to collect the following factors, which are necessary for predicting execution time and memory usage.
\begin{itemize}
    \item Execution time of each forward/ backward computation
    \item Execution time of each swapping-out/ swapping-in communication
    \item Execution timing of each swapping
    \item The sizes and order of malloc/ free operations on GPU memory
    \item Computation graph and dependency of the NN
\end{itemize}

The static prediction of all these factors is difficult. For example, the execution timing of each swapping depends on the NN structure, thus prediction in complex NNs with many branches such as GoogLeNet \cite{googlenet} is difficult. Other memory management and order of computation depend on implementation of deep learning program. 

Thus PoocH executes a runtime profiling during the first several iterations of the learning process. During the profiling execution, all feature maps are classified into \textit{swap} as the default classification. With this classification, we can collect the above information.

\subsection{Swapping-in scheduling}
\label{sec:scheduling}
\begin{figure}[tb]
    \centering
    \includegraphics[scale=0.33]{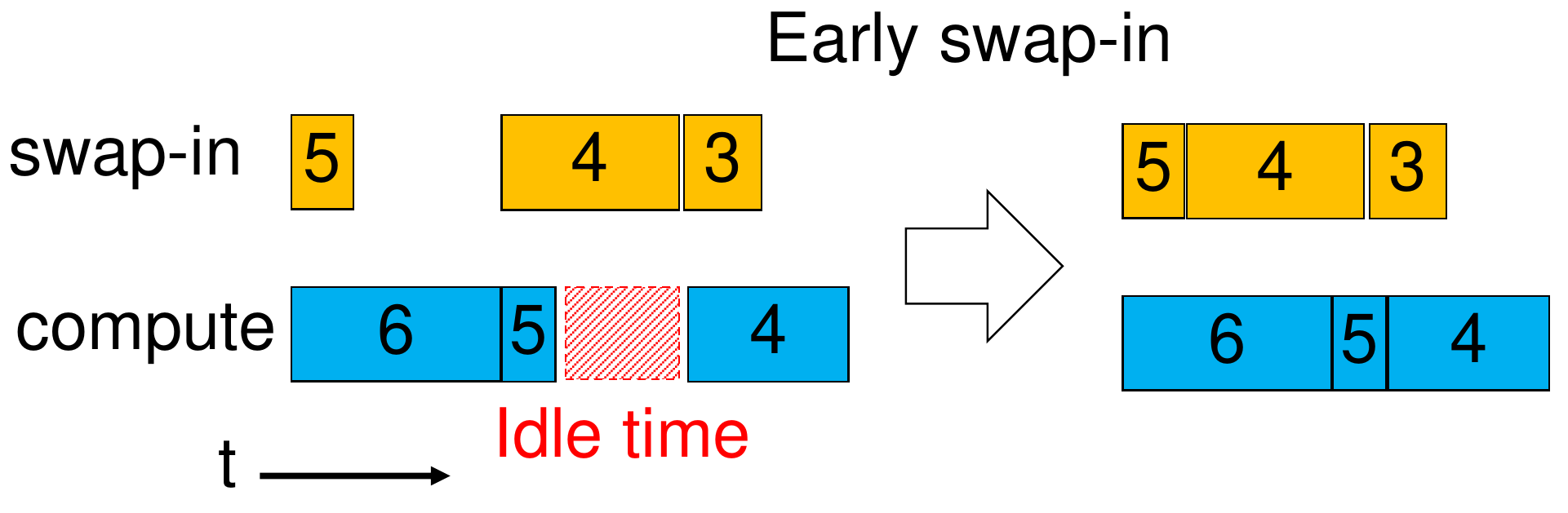}
    \caption{Example of swap-in scheduling}
    \label{fig:swapin-scheduling}
\end{figure}

This section describes improvement in pipelined swapping-in execution to reduce communication overhead. In a simple execution of backward computation with swapping (leftside in Figure \ref{fig:swapin-scheduling}), swapping-in of data for a layer (layer 4 in the figure) is overlapped with computation of the previous layer (layer 5). However, this causes unnecessary stalls. The overhead due to synchronization is reduced by moving up the schedule of swapping-in as in rightside in the figure. When this is done, GPU memory usage has to be also considered.

Through this discussion, PoocH simply executes swapping-in when there is room in the GPU memory. The amount of free memory at each time of backward can be judged from profiling result described in section \ref{sec:profiling}.

This improved scheduling is adopted both in the simulations in the classification phase and the execution phase.

\subsection{Feature maps classification}
\label{sec:search}

\subsubsection{Classification policy}
In the classification phase, PoocH classifies each feature map into one of three classes, \textit{keep}, \textit{swap}, \textit{recompute}. The purpose is to find a classification that accelerates NN computation. However, since the combination of the classification is too large ($O(3^n)$ where n denotes the number of layers in the NN.), the brute-force method may take a long time. Instead, we introduce heuristics.

The classification phase consists of the following two steps, each of which adopts heuristics for efficient search. 


\begin{enumerate}
    \item Each feature map falls into \textit{keep} or \textit{swap} tentatively (section \ref{sec:swap-search}).
    \item The targets of this phase are maps in \textit{swap} in the prior step. They are classified again into \textit{swap} or \textit{recompute}.
\end{enumerate}

In each search step, when PoocH takes a classification, it uses a simulation based on results of the runtime profiling beforehand in order to estimate the execution time and the memory usage with the classification.

\subsubsection{Classification of keep and swap}
\label{sec:swap-search}
In the first step of the classification, each feature map falls into \textit{keep} or \textit{swap}. Since the brute-force method is impractical, we describe heuristics to reduce the number of combinations to be searched.

The safest classification, which is not likely to cause out-of-memory, is one that classifies all feature maps into \textit{swap}. From this classification as the start point, we find the feature maps that cause overheads, which are maps whose swapping time cannot be hidden by computation even with pipelined execution described in Section \ref{sec:scheduling}. In this step, the search is performed as follows.

\begin{figure*}[h!]
    \centering
    \includegraphics[scale=0.33]{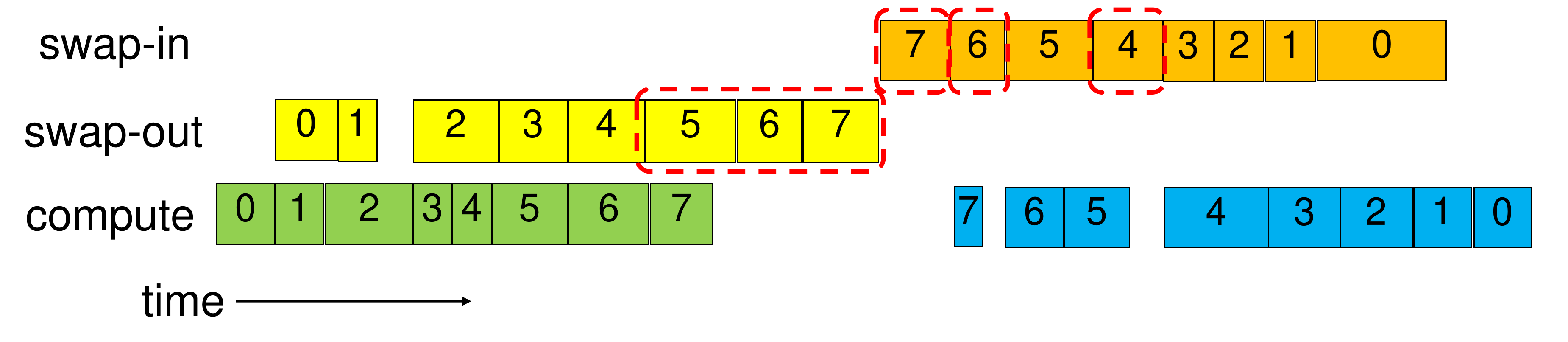}
    \caption{Example of feature maps that cause overhead by swapping. In this example, swap surrounded by the red dotted line cannot be overlapped by computation.}
    \label{fig:swap-overhead}
\end{figure*}

\begin{enumerate}
    \item The timeline in case all feature maps are swapped is simulated. Here we obtain a timeline as in Figure \ref{fig:swap-overhead}.
    \item The timeline is examined to find feature maps that do NOT cause additional overhead, as explained below. They are classified into \textit{swap} immediately in order to reduce the search space in the next step.
    \item For other (undecided in the above step) feature maps, PoocH executes a semi brute-force search in order to classify them into \textit{keep} or \textit{swap}.
\end{enumerate}

We explain examination of the timeline in the above item 2. Here we obtain a set $L_O$ of feature maps whose swapping-out time is NOT hidden by computation time. In Figure \ref{fig:swap-overhead}, $L_O = \{5,6,7\}$ (surrounded by the red dotted line). Similarly, we find a set $L_I$ by examining swapping-in, which is $L_I = \{4,6,7\}$ in the figure. For feature maps not included in $L_O$ or $L_I$ (0, 1, 2 and 3 in this example), the swapping-out/in times are hidden by computation. In our heuristics, these feature maps are classified into \textit{swap} immediately. For feature maps in $L_O \cup L_I$, classes are determined in the next search.

\begin{figure}[tb]
    \centering
    \includegraphics[scale=0.38]{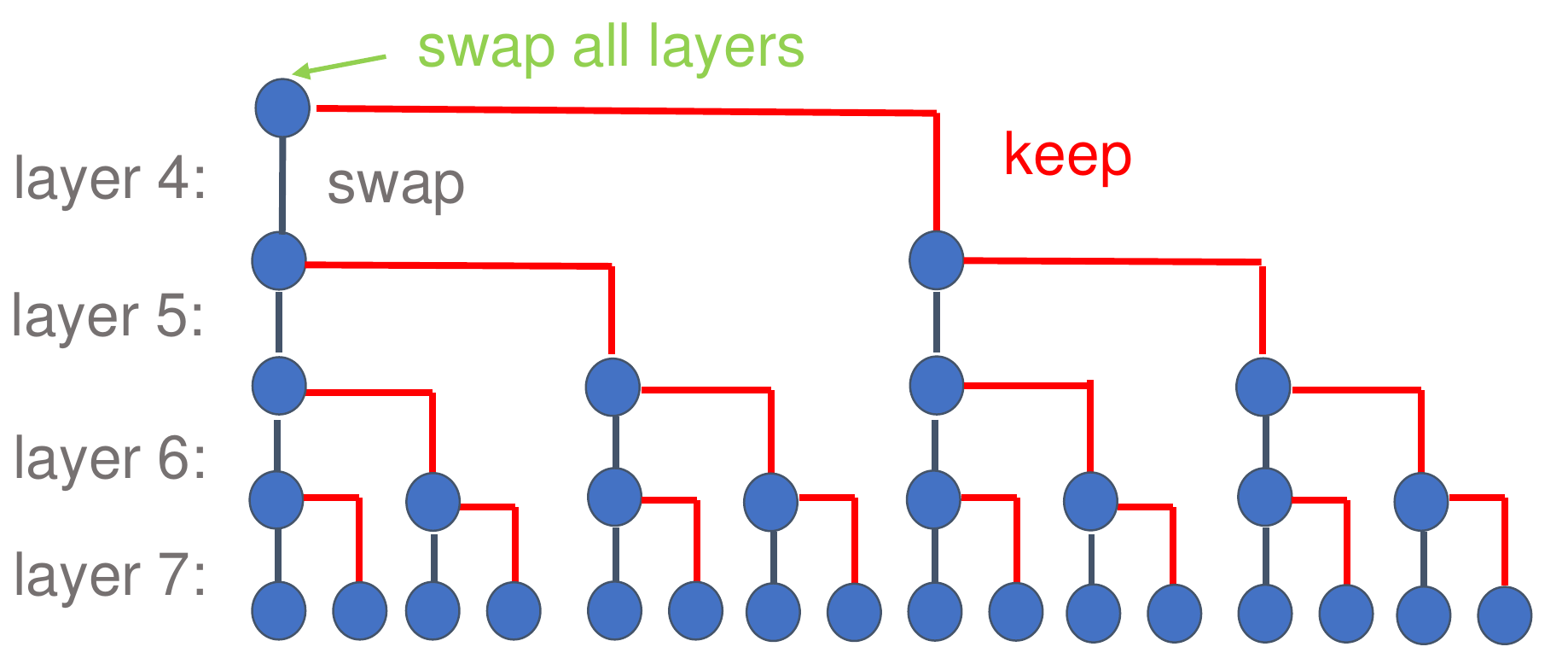}
    \caption{Search tree in decision of keep-targets and swap-targets}
    \label{fig:swap-search-tree}
\end{figure}

\subsubsection*{Reduction of search space considering swap-out}

Next, PoocH classifies feature maps in $L_O \cup L_I$ (layers 4 to 7 in the example) into \textit{keep} or \textit{swap}. We performed the brute-force search of the target feature maps, whose search tree looks like Figure \ref{fig:swap-search-tree}. However, even with a technique described above, we have found the search space of size $O(2^{|L_O\cup L_I|})$ is still too large with deep NNs. 

\begin{figure}[tb]
    \centering
    \includegraphics[scale=0.33]{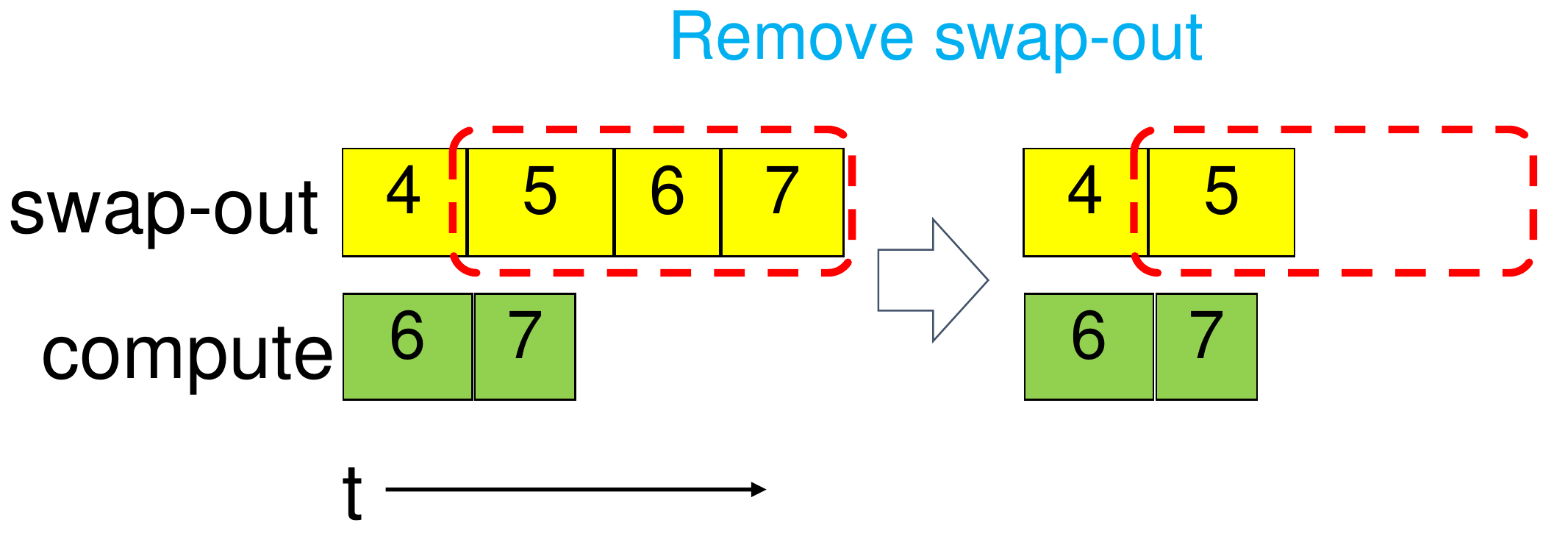}
    \caption{Reduction of swap-out. Swap-outs are reduced in order from output layer (layer 7 $\rightarrow$ 6 $\rightarrow$ 5)}
    \label{fig:remove-swapout}
\end{figure}

In order to solve this problem, we considered further reduction of the search space. For this purpose, we again investigated the execution timeline of NN computation processing in detail. Then we found that {\em swapping-out tasks which cannot be overlapped by computation tend to be continuous at the end of forward processing on the execution timeline} like the red dotted line of Figure \ref{fig:remove-swapout}. This is because (1) each forward computation does not wait swapping-out and (2) each swap-out cannot started unless the corresponding forward computation is completed. On the other hand, the tendency is much weaker in swapping-in in backward computation.

From this observation, we distinct $L_O$ and $L_I$. We adopt a light-weight greedy method for $L_O$. As shown on the right side of Figure \ref{fig:remove-swapout}, we can expect that swapping-out costs can be reduced from the back simply. In other words, those feature maps are classified into \textit{keep} as long as it can fit in the GPU memory in order from the output layer.

\begin{figure}[tb]
    \centering
    \includegraphics[scale=0.4]{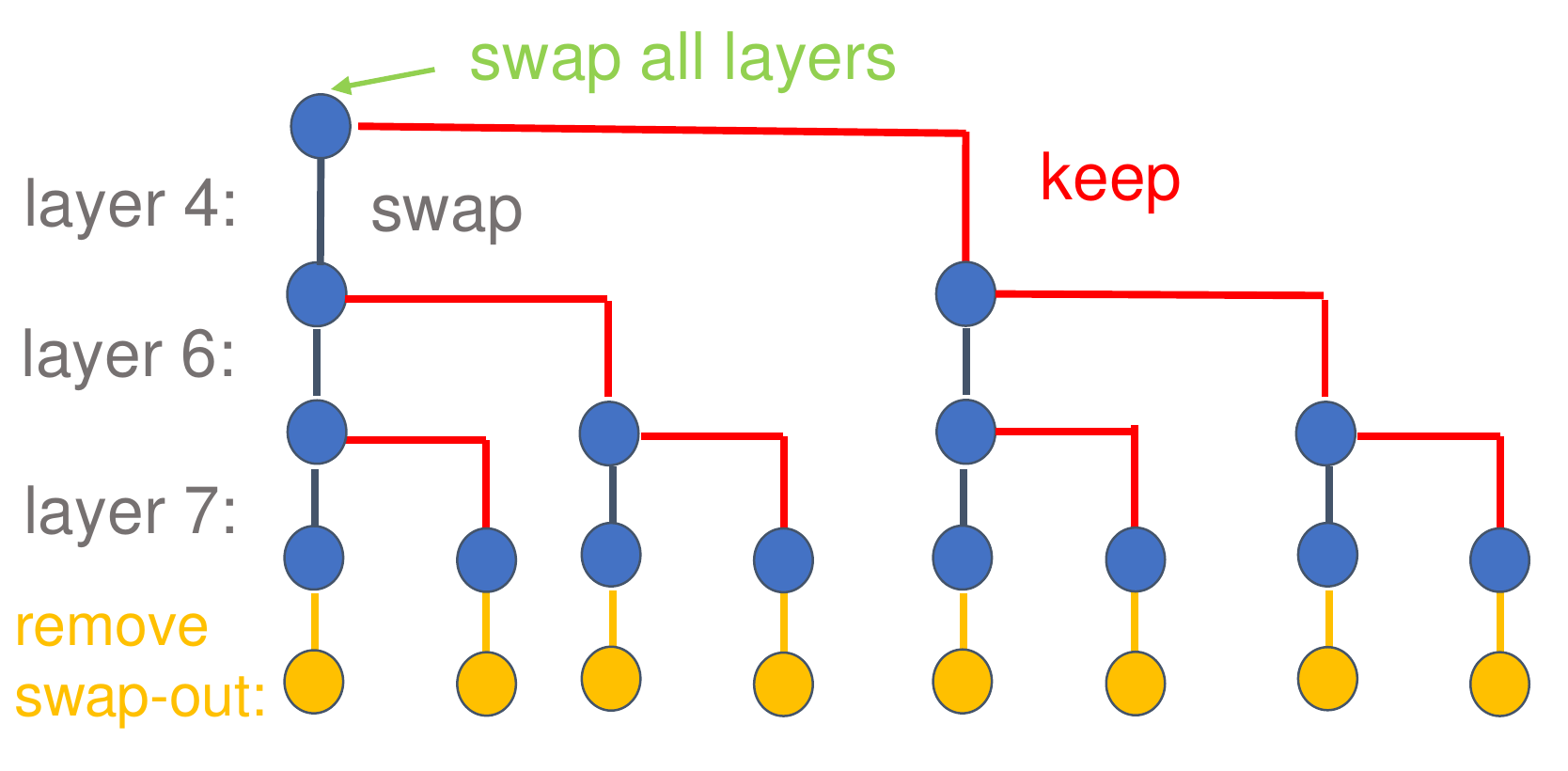}
    \caption{Search tree in decision of keep-targets and swap-targets (modified)}
    \label{fig:swap-search-tree2}
\end{figure}

On the other hand, we still use brute-force search for feature maps in $L_I$ as shown in Figure \ref{fig:swap-search-tree2}. To sum up, we execute the classification for given $L_O$ and $L_I$ as follows.

\begin{itemize}
  \item We make a binary search tree, each of whose level corresponds to an element in $L_I$.
  \item At each leaf of the search tree, feature maps in $L_I$ have been classified, while maps in $L_O \setminus L_I$ are not classified yet. We scan elements in $L_O \setminus L_I$ linearly and switch a feature map from \textit{swap} to \textit{keep}, and evaluate the entire classification by simulating the total execution time.
\end{itemize}

In the example we are using, $L_O \setminus L_I = \{5 \}$, whose elements are scanned in the later step. Generally, the search space size of the current algorithm is $O(2^{|L_I|} |L_O \setminus L_I|)$. While it still includes an exponential term, we have confirmed that it is feasible for practical deep NNs such as ResNet and ResNext as shown in Section \ref{sec:eval}.

\subsubsection{Classification of recompute}
\label{sec:recompute-search}
The targets of the second step are feature maps that have been classified into \textit{swap} in the first step tentatively. They are examined again and some feature maps are changed to \textit{recompute}. 

In this step, it is important to compare ``the overhead by data-swapping'' and ``overhead by recomputing'' for each feature map. Based on the overhead evaluation, each feature map is classified into a class that causes less overhead. The decision for feature map $X$ is done based on the following function $r(X)$.
\begin{eqnarray}
    \label{equ:recompute}
    r(X) & = & \frac{recompute\_overhead(X)}{swap\_overhead(X)}
\end{eqnarray}

Here, $swap\_overhead(X)$ is the overhead caused by swapping $X$ when classes of other feature maps are fixed (thus its value depends not only on $X$ but on others). Also, \\$recompute\_overhead(X)$ is the overhead caused by recomputing $X$.
These two functions are obtained through the timeline simulation described in Section \ref{sec:mokuteki}.

$r(X) < 1.0$ means that the overhead for $X$ is smaller with recomputing than with swapping. Especially, as $r(X)$ is smaller, more overhead is likely to be reduced by switching from \textit{swap} to \textit{recompute}. On the other hand, if $r(X) > 1.0$, \textit{swap} is better. Considering the above, this step is performed in the following flow.
\begin{enumerate}
    \item We let be $L$ the set of feature maps classified into \textit{swap} in the first step of search.
    \item $r(X)$ is evaluated for each feature map $X$ in $L$.
    \item All $X$ that satisfy $r(X) \geq 1.0$ are removed from $L$, and classified into \textit{swap}.
    \item We pick up an $X$ that satisfies ``$r(X) < 1.0$'' and ``$r(X)$ is the smallest''. We remove $X$ from $L$, and classify into \textit{recompute}.
    \item If $ L $ is empty, the search is terminated. Otherwise we return to (2).
\end{enumerate}

The compute complexity of this step is $O(|L|^2)$, which is much smaller than that of step 1.

\section{EVALUATION} 
\label{sec:eval}
We have implemented PoocH as an extension to Chainer version 3, a deep learning framework developed by PFN\cite{chainer}. Chainer implicitly supports computation on GPU by calling its own CUDA kernel and NVIDIA's cuDNN library APIs \cite{cudnn}. However, the original Chainer fails the execution with large NNs that require more memory than GPU memory capacity.

We have implemented the runtime profiling mechanism and the classification algorithm of PoocH inside Chainer. Especially, GPU memory management functions (allocate and free) of Chainer are hooked in order to keep track of GPU memory usage in the profiling phase.

This section shows performance evaluation results of this extended Chainer.
As execution environments, we used two machines, an Intel Xeon based machine (x86 machine) and a POWER9 machine. In each machine, a single Tesla V100 GPU with 16 GB memory is used. The details of evaluation environments are shown in the Table \ref{tab:env} and Table \ref{tab:env2}. Between the two machines, it is important that CPU-GPU interconnects are different. The x86 machine adopts ordinary PCI-Express gen3 link and the POWER9 machine adopts NVLink2.0, which is more than four times faster than PCI-Express gen3. 

\begin{table}[tb]
    \centering
    \caption{The evaluation environment (x86 machine)}
    \begin{tabular}{c|c} \hline \hline
        GPU & NVIDIA Tesla V100 \\ 
        GPU memory capacity & 16 GB \\ \hline
        CPU &  Intel Xeon Gold 6140  \\ 
        CPU memory capacity & 192 GB \\ \hline
        CPU-GPU interconnect &  PCIe gen3 x16 \\ 
        CPU-GPU bandwidth &  16 GB/sec \\ \hline
        OS & CentOS 7.4 \\
        CUDA &  CUDA 9.1 \\ 
        cuDNN &  cuDNN 7.1 \\ \hline
    \end{tabular}
    \label{tab:env}
\end{table}

\begin{table}[tb]
    \centering
    \caption{The evaluation environment (IBM POWER9 machine)}
    \begin{tabular}{c|c} \hline \hline
        GPU & NVIDIA Tesla V100 \\ 
        GPU memory capacity & 16 GB \\ \hline
        CPU & IBM POWER9 \\ 
        CPU memory capacity & 1 TB \\ \hline
        CPU-GPU interconnect &  NVLink2.0 x2 \\ 
        CPU-GPU bandwidth &  75 GB/sec \\ \hline
        OS & RHEL 7.5 (Maipo) \\
        CUDA &  CUDA 9.2 \\ 
        cuDNN &  cuDNN 7.1 \\ \hline
    \end{tabular}
    \label{tab:env2}
\end{table}

\subsection{Evaluation of each optimization}
As described in chapter \ref{sec:pooch}, PoocH has several optimization steps, swapping timing, classification into swap/keep, and classification into recompute. We evaluate the impact on performance by each optimization by comparing the following four methods. 
\begin{itemize}
    \item swap-all(w/o scheduling) : All feature maps are swapped. Each swapping-in simply starts simultaneously with the previous computation.
    \item swap-all : All feature maps are swapped. The timing of swapping-in is improved as described in Section \ref{sec:scheduling}.
    \item swap-opt : Feature maps are classified into \textit{keep} or \textit{swap} by conducting only Step 1 in Section \ref{sec:swap-search}.
    \item PoocH : Feature maps are classified into \textit{keep}, \textit{swap}, \textit{recompute} by conducting Step 1 and Step 2 in Section \ref{sec:search}.
\end{itemize}

``Swap-opt'' and ``PoocH'' also uses optimization of the timing of swapping.

\begin{figure}[tb]
    \centering
    \includegraphics[scale=0.42]{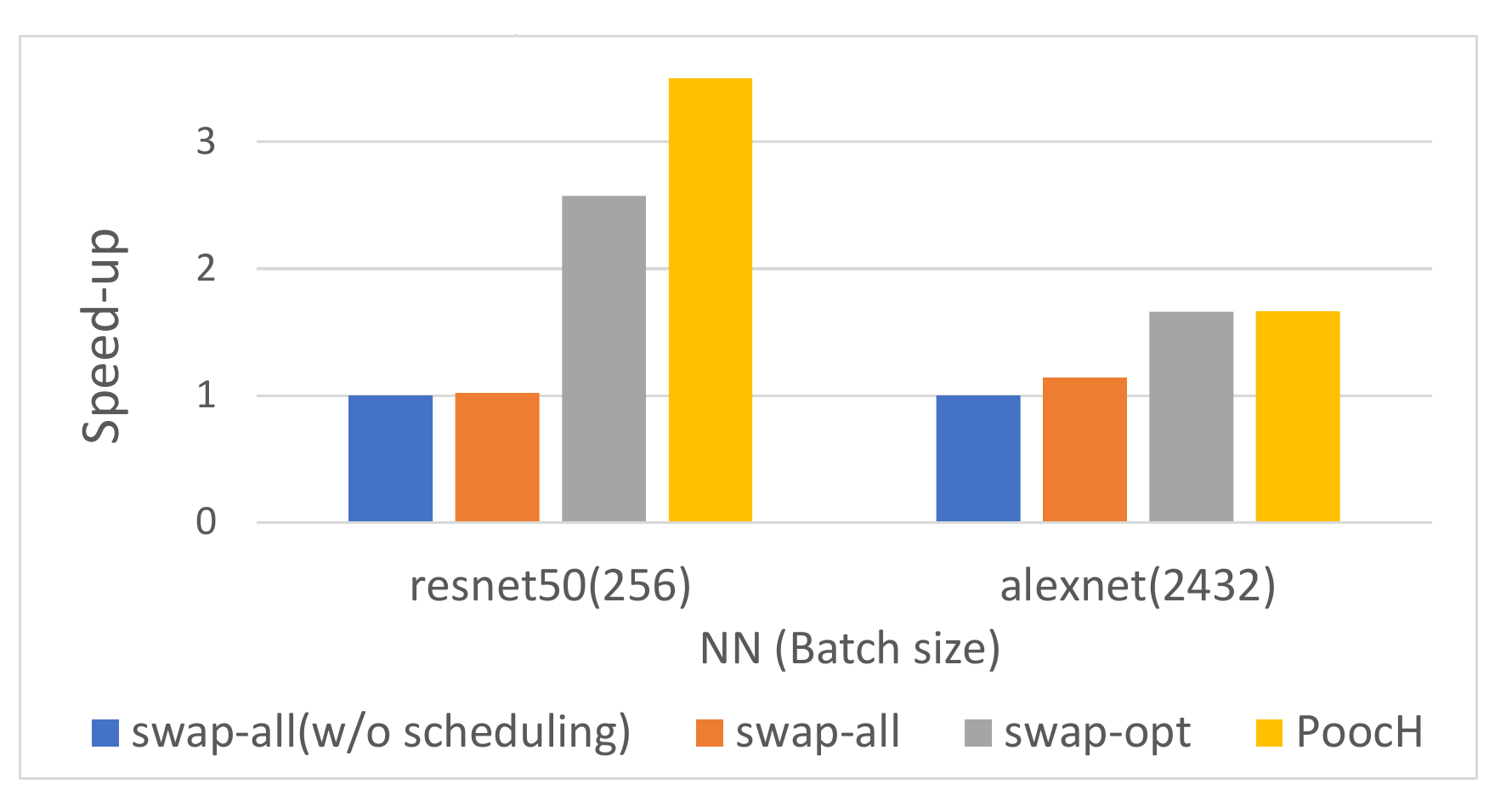}
    \caption{Evaluation of each optimization on x86 machine (speed up for swap-all (w/o scheduling))}
    \label{fig:test1_each_optimize}
\end{figure}

Figure \ref{fig:test1_each_optimize} shows the comparison results on x86 machine. In the experiment, we used ResNet50 \cite{resnet}, and AlexNet \cite{alexnet}. All three are NNs for image recognition and the batch sizes are configured so that problem sizes exceed the GPU memory capacity. In this experiment, the speeds are given by batch size divided by execution time per learning iteration $[\#images/s]$. The graph shows relative speed-up where the base case is ``swap-all(w/o scheduling)''. 

In Figure \ref{fig:test1_each_optimize}, we observe that ``swap-all'' improves the performance by 2-14 \% compared to swap-all (w/o scheduling). This is because the execution timing of each swap-in is moved up so that the CPU-GPU communications became likely to be overlapped by computation. The performance of ``Swap-opt'' is improved further by keeping some data on GPU memory and by reducing the communication amounts. As a result, the performance is improved by x1.4 to x3.0 compared with ``swap-all''.

For all of three NNs, PoocH with all optimizations shows the highest performance. Especially the PoocH's performance was x1.45 swap-opt for ResNet50. We consider this large difference comes from the characteristics of ResNet50. ResNet50 has many layers with small computation complexity with respect to the feature map sizes. Therefore, recomputing such feature maps tends to cause less overhead than swapping. On the other hand, for NNs with large computational cost like AlexNet, recomputation was rarely chosen because there is computation time enough to overlap swap. Thus, the performance difference between PoocH and swap-opt for Alexnet is small.

\begin{figure}[tb]
    \centering
    \includegraphics[scale=0.42]{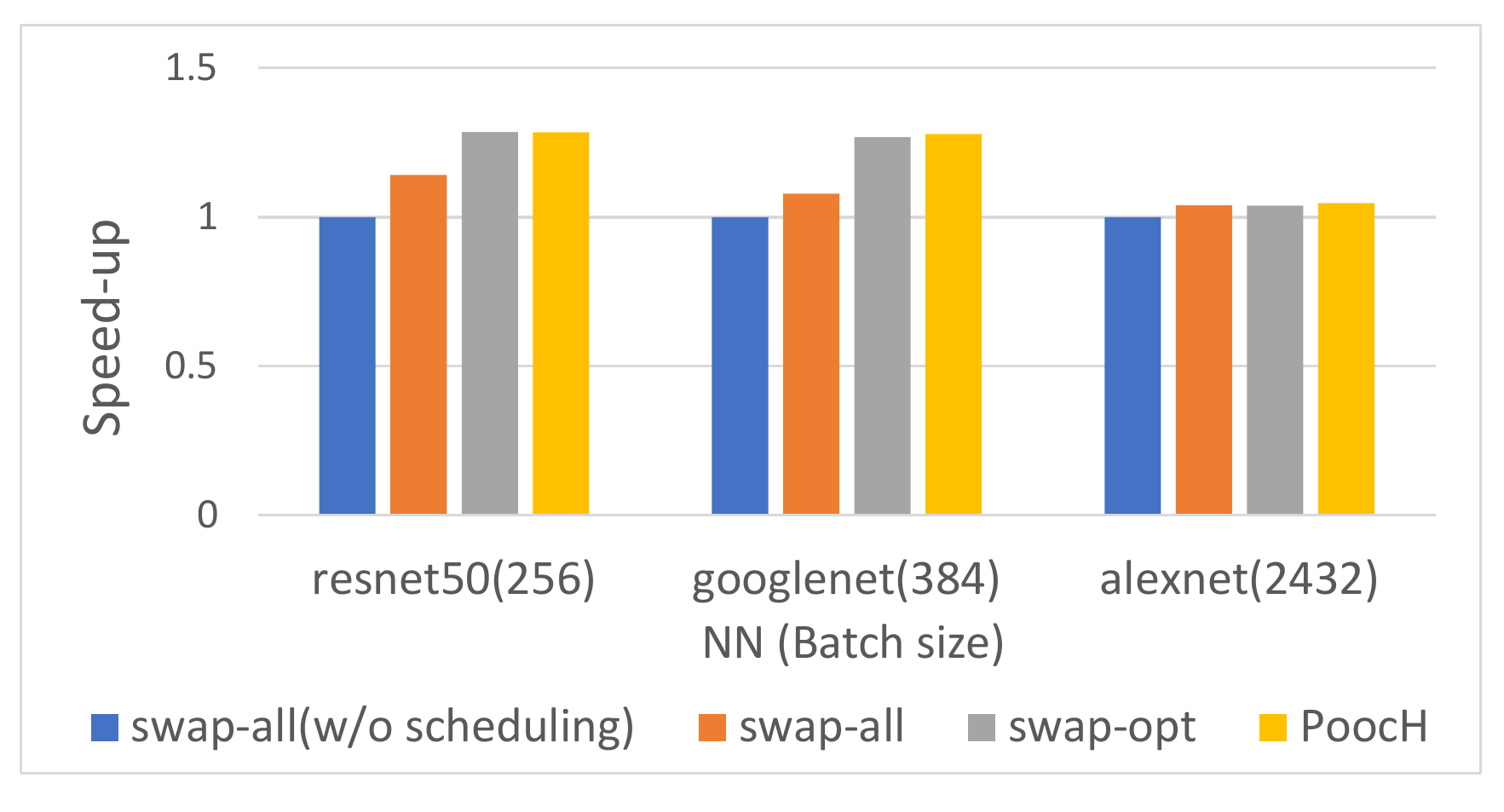}
    \caption{Evaluation of each optimization on POWER9 machine (speed up for swap-all (w/o scheduling))}
    \label{fig:test1_each_optimize_power}
\end{figure}

Figure \ref{fig:test1_each_optimize_power} shows the results on POWER9 machine. We observe PoocH showed the best performance also on this machine. In the figure, we observe the differences between ``swap-opt'' and PoocH are small compared to those in Figure \ref{fig:test1_each_optimize}. This is because the overhead of data-swapping is originally small since NVLink accelerates CPU-GPU communication on POWER9 machine. In addition, thanks to high speed NVLink, swapping communications can be sufficiently overlapped by computation even for NNs with a small computation complexity such as ResNet50.

\subsection{Comparison with existing methods}
\label{sec:NN_test}
In this experiment, we computed ResNet50 and AlexNet with various batch sizes. In addition, we computed ResNext101 (3D) \cite{resnext} for various input data sizes with batch size of 1. ResNext101 (3D) is an extension of ResNext101 for video recognition based on \cite{3dnn}. The input data of this NN is 3D data, and the required memory may exceed GPU memory capacity even with batch size of 1. 

The computation of each NN was performed by the following three methods, respectively.
\begin{itemize}
    \item in-core : Neither data-swapping nor recomputing is performed.
    \item superneurons : The classification based on SuperNeurons \cite{superneurons} is used.
    \item PoocH : Our proposed system based on runtime profiling is used.
\end{itemize}

SuperNeurons uses the following hybrid classification algorithm. 
\begin{itemize}
    \item Feature maps are stored on GPU memory preferentially from output layer.
    \item Among the feature maps that do not fit in the GPU memory, the feature maps of convolution layer are targets of swapping. The feature maps of layers with other types are recomputed.
    \item Each swap-in starts simultaneously with the computation of the immediately preceding convolution layer.
\end{itemize}
For the measurement, we have implemented this algorithm by extending Chainer. 

In this experiment, the experimental results do not include profiling and optimization time of PoocH. Although it gets longer with the number of layers, we observed that it was about 2 minutes even for resnext101 with >300 layers. Thus, it can be easily amortized by speed-up of learning when the whole learning takes hours or days.

\begin{figure}[tb]
    \centering
    \includegraphics[scale=0.32]{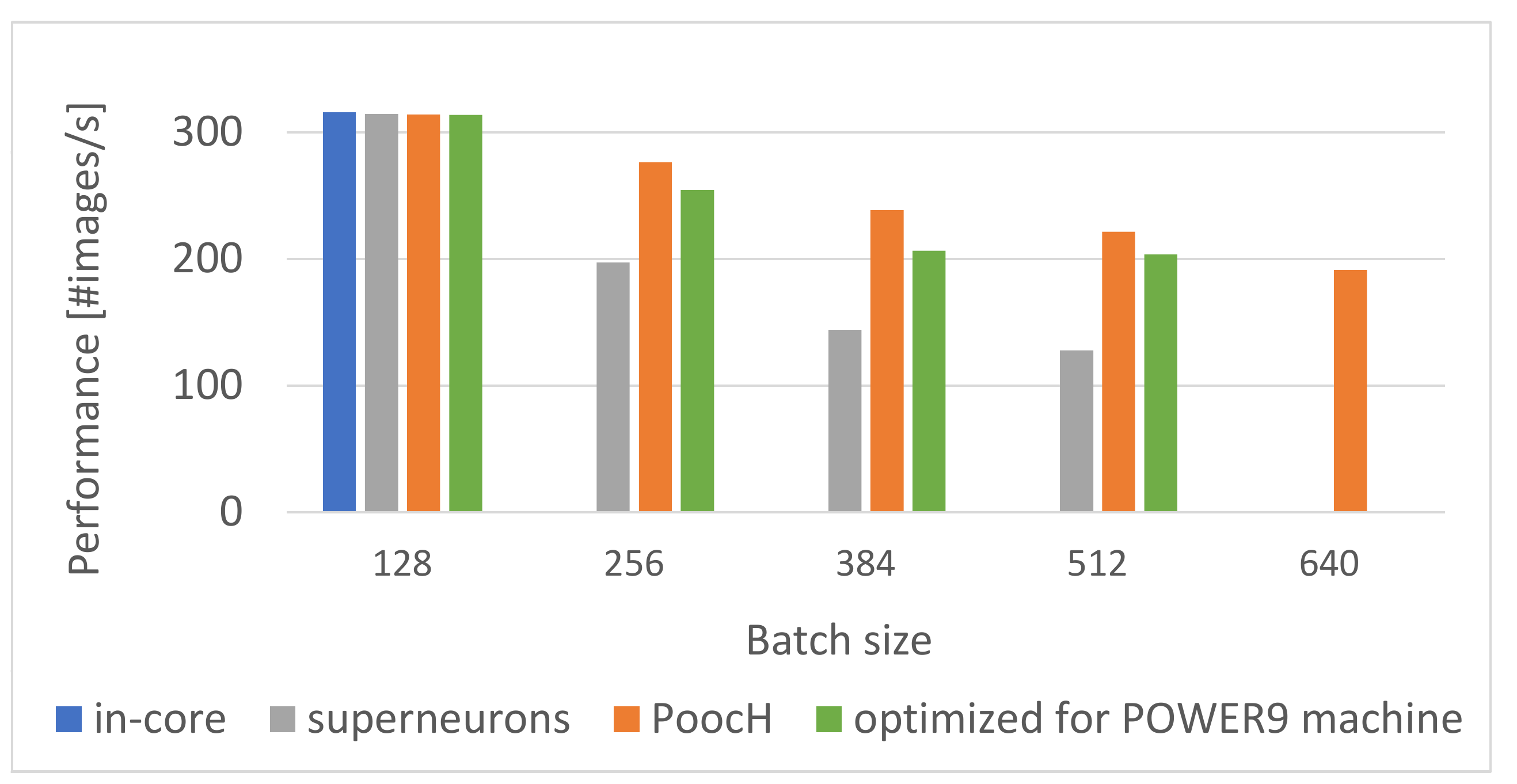}
    \caption{Performance for ResNet50 on x86 machine}
    \label{fig:test2_resnet}
\end{figure}

\begin{figure}[tb]
    \centering
    \includegraphics[scale=0.25]{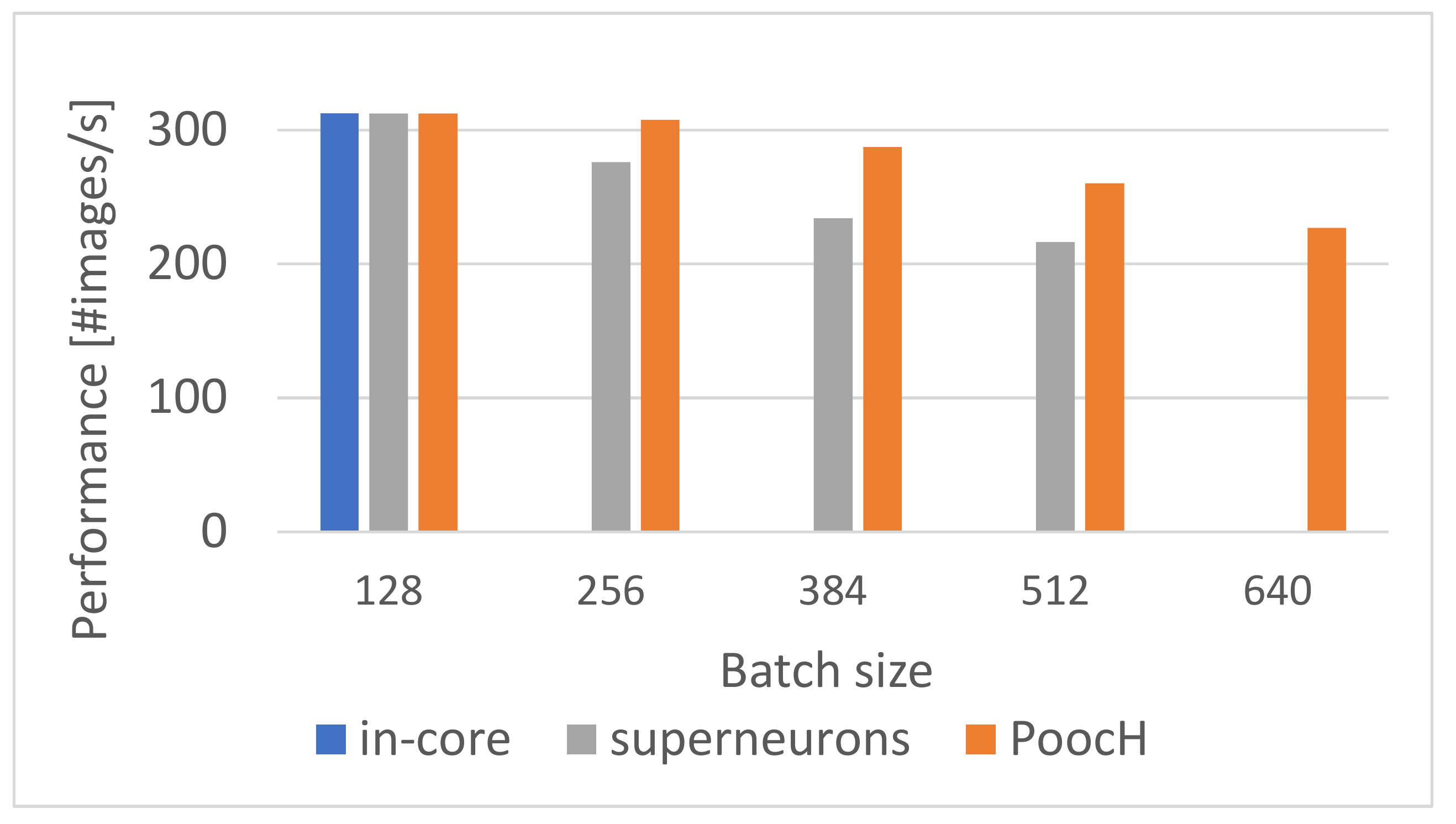}
    \caption{Performance for ResNet50 on POWER9 machine}
    \label{fig:test2_resnet_power}
\end{figure}

The results for ResNet50 are shown in Figures \ref{fig:test2_resnet} and \ref{fig:test2_resnet_power}. Each figure corresponds to results on the x86 machine and the POWER9 machine, respectively. We observe that when the batch size is set to $256$ or more, the memory usage exceeded the GPU memory capacity and ``in-core'' execution fails. When batch size is $640$, the memory usage was more than 50 GB. PoocH successfully computes this case.

In Figure \ref{fig:test2_resnet}, the performance of in-core was $316$ [\#images/s], and the performance of PoocH was 195 to 316 [\#images/s]. For the problem size exceeding the GPU memory capacity, PoocH can perform computation with $13-38 \%$ performance degradation compared with ``in-core'' case. 

Both in PoocH and superneurons, performances are degraded as the batch size increases. This is due to an increase of feature maps that cannot be stored on the GPU memory as batch size increases. For larger cases, we need to make more feature maps swapped or recomputed, which leads performance degradation.

However, comparing the two, PoocH shows x1.40 - x1.73 better performances in the range of batch size 256 to 512. With PoocH, targets of swapping and recomputation are determined by considering actual each layer's computation time and each swap communication time, not the type of layer. As a result, performance was improved compared with static method like superneurons.

Besides, execution of superneurons fails with batch size of 640. This is because superneurons schedules swapping-in without considering the actual memory usage, resulting in GPU memory shortage. With PoocH, such problems do not occur since swap-in communications are scheduled considering memory usage based on profiling.

Figure \ref{fig:test2_resnet_power} shows the performance on a POWER9 machine. Here performance degradation of PoocH compared with in-core was $2-28 \%$. On this machine, NVLink accelerates CPU-GPU communication, so the overhead of data-swapping is small. As a result, performance degradation in the case of POWER9 was smaller than in the case of x86.

We also note that PoocH can capture the differences in characteristics of the two machines. Table \ref{tab:optimize_result} shows the number of feature maps classified as \textit{keep}, \textit{swap} or \textit{recompute} for ResNet50 when using PoocH or superneurons on both environments. From the table, we confirmed that PoocH classifies more feature maps into recompute on the x86 machine than on the POWER9 machine. This is because PoocH tends to prefer recomputing in the x86 machine with slower interconnect. On the other hand, superneurons makes the same classification on the two machines. 

In addition, we computed ResNet50 on x86 machine with the PoocH's optimization results for POWER9 machine (Figure \ref{fig:test2_resnet}). In Figure \ref{fig:test2_resnet}, using optimization results for x86 machine resulted in better performance than using optimization results for POWER9 machine. Besides, execution using optimization results for POWER9 machine fails with batch size of 640 on x86 machine. The order of malloc/free are different since the execution time of computation/data-swapping are different on both environments. Therefore, using the classification according to the different environment may cause GPU memory shortage. 

As above, PoocH successfully optimizes the classification according to the execution environment and shows better performance on both machines.

\begin{table}[tb]
    \centering
    \caption{The number of feature maps classified as \textit{keep}, \textit{swap} or \textit{recompute} for ResNet50 (batch size = 512) when using PoocH or superneurons (on both environments)}
{\small
    \begin{tabular}{c|c c c} \hline \hline
         & \#\textit{keep} & \#\textit{swap} & \#\textit{recomp} \\ \hline
        PoocH (x86) & 66 & 12 & 27 \\ 
        superneurons (x86) & 66 & 21 & 18 \\ \hline
        PoocH (POWER9) & 66 & 36 & 3 \\ 
        superneurons (POWER9) & 66 & 21 & 18 \\ \hline
    \end{tabular}
}
    \label{tab:optimize_result}
\end{table}

\begin{figure}[tb]
    \centering
    \includegraphics[scale=0.36]{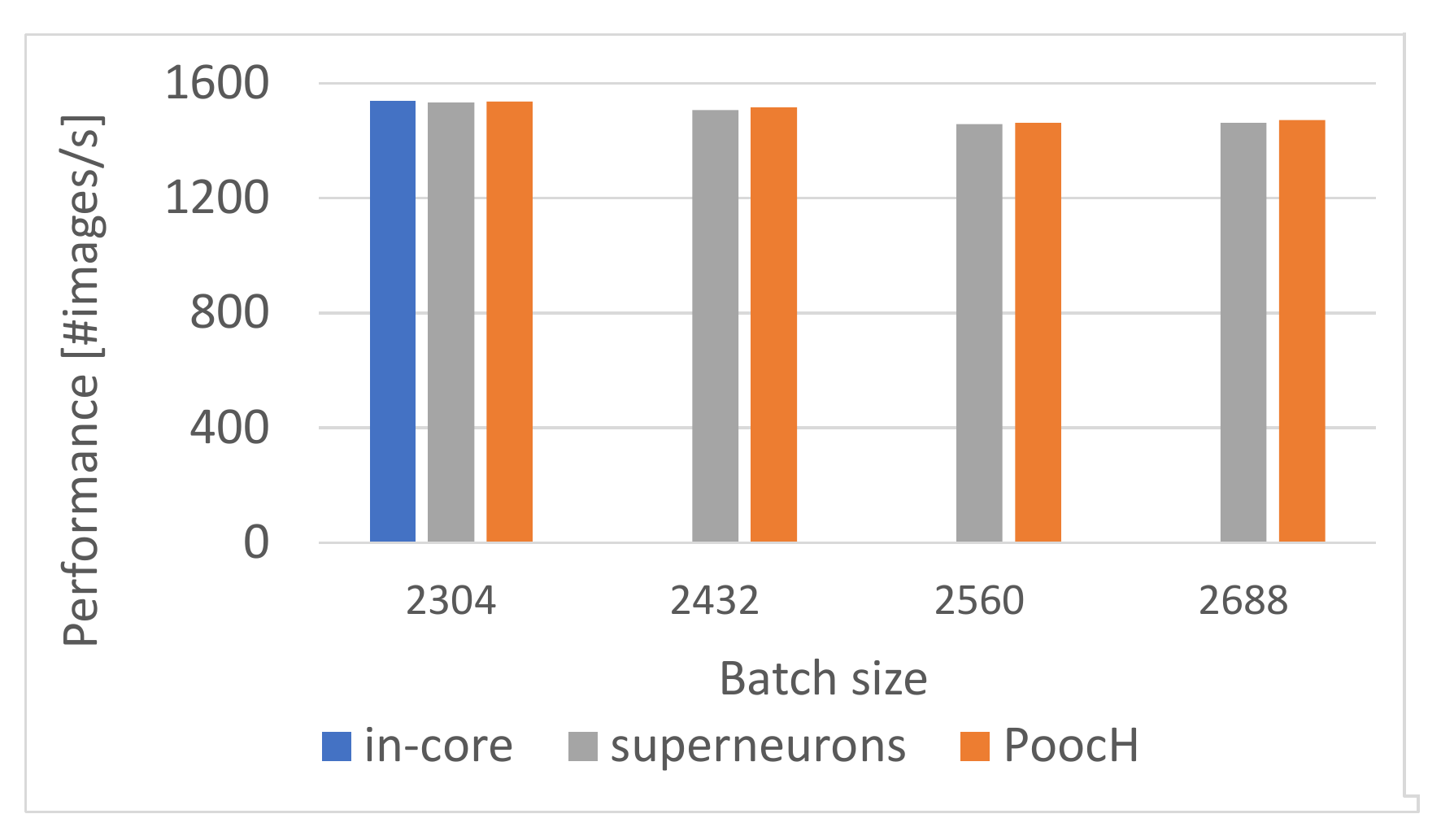}
    \caption{Performance for AlexNet on x86 machine}
    \label{fig:test2_alexnet}
\end{figure}

\begin{figure}[tb]
    \centering
    \includegraphics[scale=0.36]{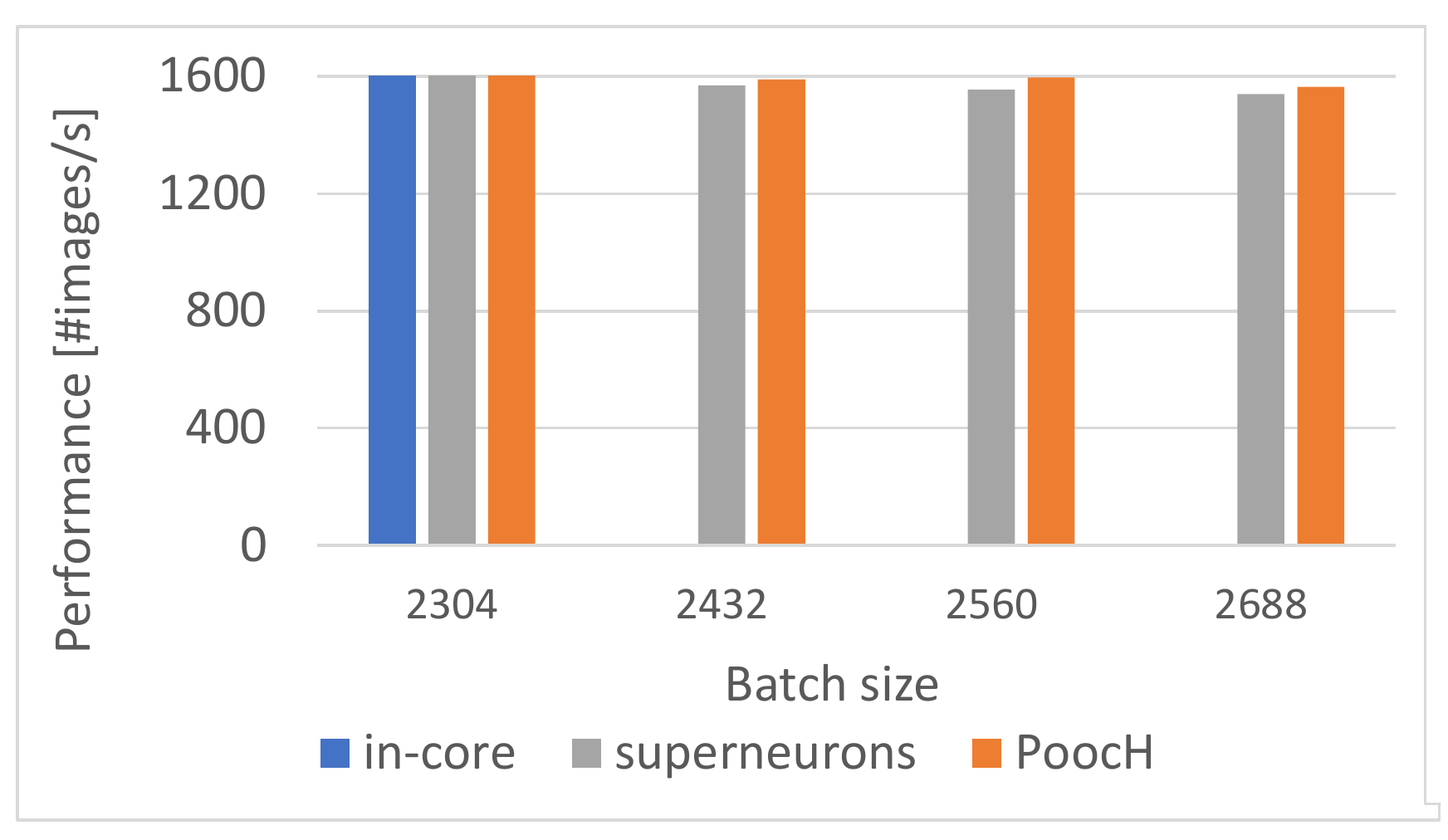}
    \caption{Performance for AlexNet on POWER9 machine}
    \label{fig:test2_alexnet_power}
\end{figure}

The results for AlexNet are shown in Figures \ref{fig:test2_alexnet} and \ref{fig:test2_alexnet_power}. In both figures, the performance degradation of PoocH was less than $6.1 \% $ compared with in-core. This is because layers in AlexNet have large computation complexities, so recomputation is rarely done when using PoocH, and swap communication could be overlapped sufficiently. Also, the performance difference between PoocH and superneurons was small.

\begin{figure}[tb]
    \centering
    \includegraphics[scale=0.25]{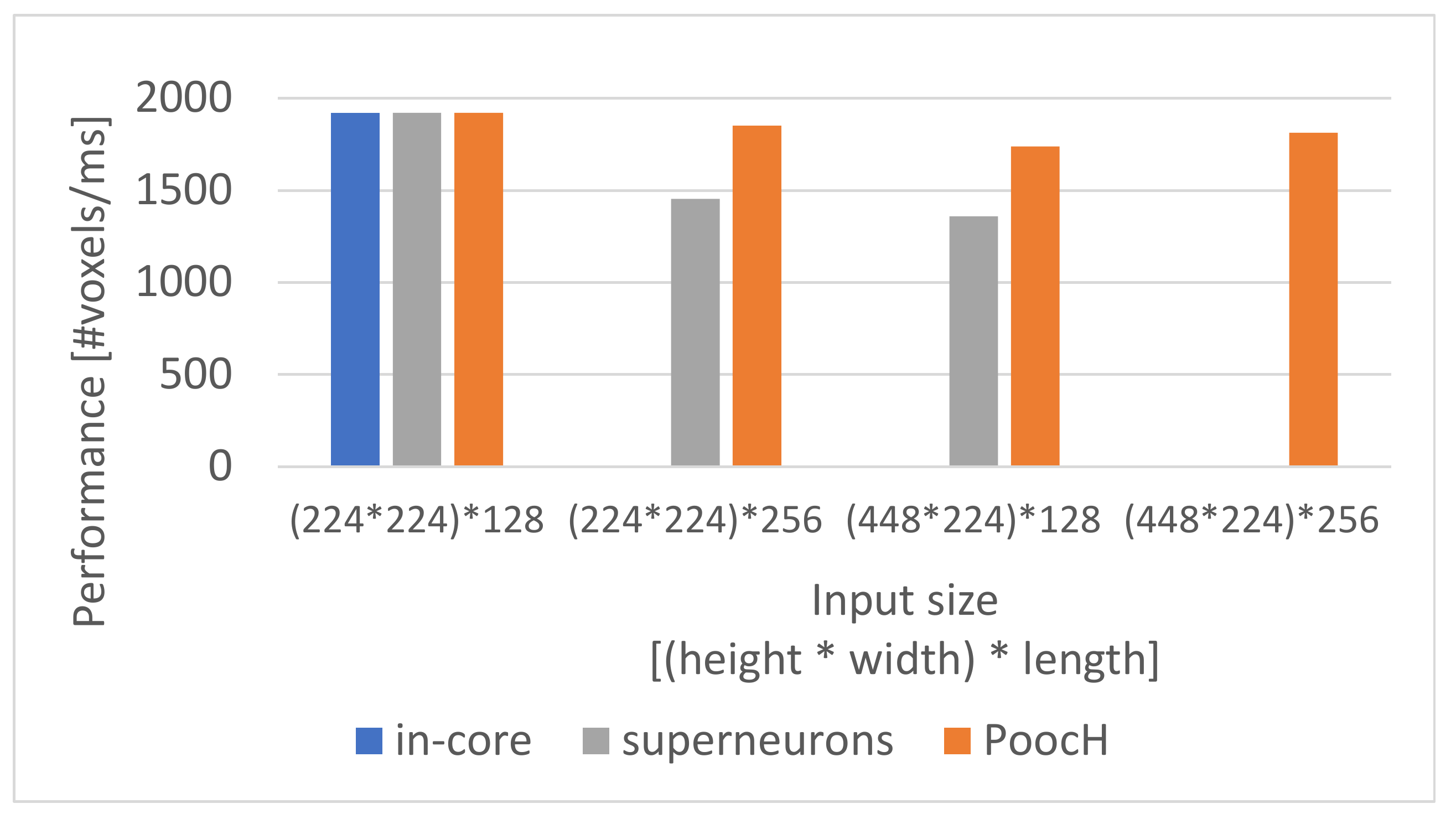}
    \caption{Performance for ResNext101 (3D) on x86 machine}
    \label{fig:test2_resnext_3d}
\end{figure}

\begin{figure}[tb]
    \centering
    \includegraphics[scale=0.25]{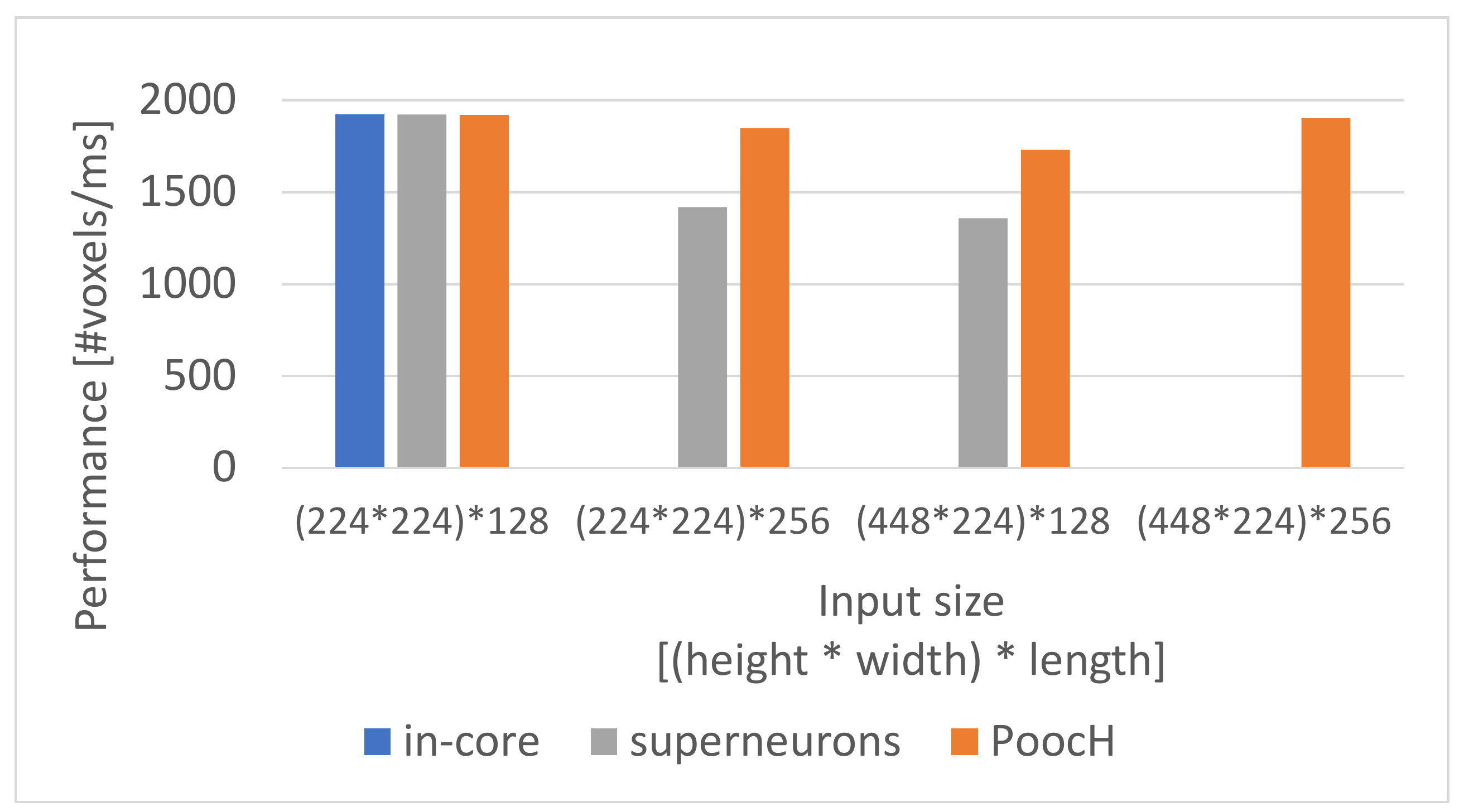}
    \caption{Performance for ResNext101 (3D) on POWER9 machine}
    \label{fig:test2_resnext_3d_power}
\end{figure}

Finally, the results for ResNext101 (3D) are shown in Figures \ref{fig:test2_resnext_3d} and \ref{fig:test2_resnext_3d_power}. With this NN, the memory consumption depends on the sizes of 3D input data. When we use large images as input, the memory requirement exceeds GPU memory capacity even with batch size of 1. PoocH successfully compute such cases. 

For ResNext101 (3D), the performance degradation of PoocH compared with ``in-core'' case was less than $10 \% $ on both environments. Computation of NNs for 3D data tends to have a large computation complexity, thus swap communication was sufficiently overlapped by computation. PoocH also had better performance compared to superneurons.

We have shown that PoocH improves the computation performance of large NNs. It is based on optimizing according to the structure of the NN, the data size and the execution environment.

\section{RELATED WORK}
As described in section \ref{sec:hybrid},  SuperNeurons has been proposed by Wang et al \cite{superneurons}. SuperNeurons uses the hybrid method of data-swapping and recomputing. Their method statically decides classification unlike PoocH. We have shown that PoocH enables faster executions using classification algorithm based on runtime profiling.

Rhu et al. proposed vDNN to compute large scale NNs using data-swapping method \cite{vdnn}. Although vDNN optimize classification of swapping, it does not incorporate the recomputing method. Also, Cho et al. implemented the data-swapping method on Chainer \cite{oocchainer}, and our Chainer is an extension of theirs. Ito's ooc\_cuDNN library performs data-swapping after dividing each computation and data \cite{ooccudnn}. \\ooc\_cuDNN supports NNs where memory consumption of a single layer exceeds the GPU memory capacity. By integrating PoocH and ooc\_cuDNN, PoocH will support NNs of wider ranges.

Chen et al. proposed to compute large-scale NNs by using recomputing method \cite{gradientcheckpoint}. PoocH combines recomputing method with data-swapping method.

\section{SUMMARY AND FUTURE WORK}
This paper described PoocH that supports efficient execution of large neural networks that require more memory then GPU memory capacity. It reduces the performance overhead by determining target layers of swapping or recomputing based on runtime profiling. In addition, PoocH schedules swapping-in efficiently. By using PoocH, the performance degradation when computing NNs exceeding the GPU memory capacity is reduced to $38 \%$ on x86 machine and $28 \%$ on POWER9 machine.

The current version of PoocH targets only NNs that compute the same problem size in each learning iteration. As future work, we will extend PoocH in order to deal with NNs whose problem sizes change for each iteration.

\bibliographystyle{plain}
\bibliography{ito-pooch}

\end{document}